\documentclass[runningheads]{llncs}

\pdfoutput=1
\usepackage{grffile}

\usepackage{graphicx}
\usepackage{comment}
\usepackage{amsmath,amssymb} %
\usepackage{color}
\usepackage{comment}
\usepackage{wrapfig}
\usepackage{multirow}

\usepackage{arydshln}

\newcommand{\etal}{\textit{et al.}}
\DeclareMathOperator*{\argmin}{argmin}
\begin{document}
\pagestyle{headings}
\mainmatter
\def\ECCVSubNumber{2547}  %

\title{PatchNets: Patch-Based Generalizable Deep Implicit 3D Shape Representations} %

\titlerunning{PatchNets}
\author{Edgar Tretschk\inst{1}%
\hspace{27pt}
Ayush Tewari\inst{1}%
\hspace{24pt}
Vladislav Golyanik\inst{1}%
\\
\hspace{-7pt}
Michael Zollh\"ofer\inst{2}
\hspace{19pt}
Carsten Stoll\inst{2}
\hspace{23pt}
Christian Theobalt\inst{1}
} 
\authorrunning{E. Tretschk et al.}
\institute{Max Planck Institute for Informatics, Saarland Informatics Campus%
\and
Facebook Reality Labs
}
\maketitle

\setcounter{footnote}{0}

\begin{abstract}
Implicit surface representations, such as signed-distance functions,  combined with deep learning have led to impressive models which can  represent detailed shapes of objects with arbitrary topology. 
Since a continuous function is learned, the reconstructions can also be extracted at any arbitrary resolution. 
However, large datasets such as ShapeNet are required to train such models.

In this paper, we present a new mid-level patch-based surface representation.
At the level of patches, objects across different categories share similarities, which leads to more generalizable models.
We then introduce a novel method to learn this patch-based representation in a canonical space, such that it is as object-agnostic as possible. 
We show that our representation trained on one category of objects from ShapeNet can also well represent detailed shapes from any other category. 
In addition, it can be trained using much fewer shapes, compared to existing approaches. 
We show several applications of our new representation, including shape interpolation and partial point cloud completion. 
Due to explicit control over positions, orientations and scales of patches,
our representation is also more controllable compared to object-level representations, which enables us to deform encoded shapes non-rigidly. 


\end{abstract}

\section{Introduction}
Several 3D shape representations exist in the computer vision and computer graphics communities, such as point clouds, meshes, voxel grids and implicit functions. 
Learning-based approaches have mostly focused on voxel grids due to their regular structure, suited for convolutions. 
However, voxel grids \cite{Choy2016} come with large memory costs, limiting the output resolution of such methods. 
Point cloud based approaches have also been explored \cite{qi2017pointnet++}.
While most approaches assume a fixed number of points, recent methods also allow for variable resolution outputs \cite{sitzmann2019scene,mescheder2019occupancy}. 
Point clouds only offer a sparse representation of the surface. 
Meshes with fixed topology are commonly used in constrained settings with known object categories \cite{Wang2018}. 
However, they are not suitable for representing objects with varying topology.
Very recently, implicit function-based representations were introduced~\cite{park2019deepsdf,mescheder2019occupancy,chen2019learning}.
DeepSDF~\cite{park2019deepsdf} learns a network which represents the continuous signed distance functions for a class of objects. 
The surface is represented as the $0$-isosurface. 
Similar approaches~\cite{mescheder2019occupancy,chen2019learning} use occupancy networks, where only the occupancy values are learned (similar to voxel grid-based approaches), but in a continuous representation.
Implicit functions allow for representing (closed) shapes of arbitrary topology.
The reconstructed surface can be extracted at any resolution, since a continuous function is learned.

\begin{wrapfigure}{R}{0.38\textwidth} 
\centering 
\includegraphics[width=0.38\textwidth]{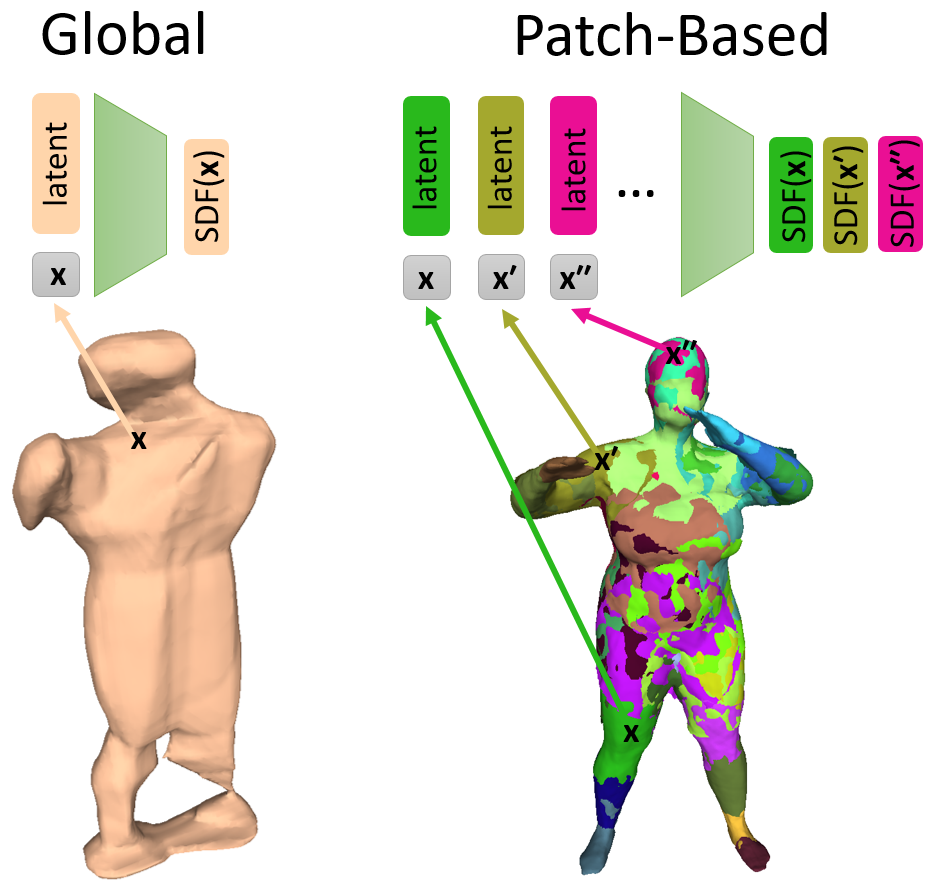} 
\caption{In contrast to a global approach, our patch-based method generalizes to human shapes after being trained on rigid ShapeNet objects.} 
\label{fig:teaser} 
\end{wrapfigure} 

All existing implicit function-based methods rely on large datasets of 3D shapes for training. 
Our goal is to build a generalizable surface representation which can be trained with much fewer shapes, and can also generalize to different object categories.
Instead of learning an object-level representation, our \emph{PatchNet} learns a mid-level representation of surfaces, at the level of patches. 
At the level of patches, objects across different categories share similarities.
We learn these patches in a canonical space to further abstract from object-specific details. 
Patch extrinsics (position, scale and orientation of a patch) allow each patch to be translated, rotated and scaled. 
Multiple patches can be combined in order to represent the full surface of an object. 
We show that our patches can be learned using very few shapes, and can  generalize across different object categories, see Fig.~\ref{fig:teaser}. 
Our representation also allows to build object-level models, \emph{ObjectNets}, which is useful for applications which require an object-level prior. 

We demonstrate several applications of our trained models, including partial point cloud completion from depth maps, shape interpolation, and a generative model for objects.
While implicit function-based approaches can reconstruct high-quality and detailed shapes, they lack controllability. 
We show that our patch-based implicit representation \emph{natively} allows for controllability due to the explicit control over patch extrinsics.
By user-guided rigging of the patches to the surface, we allow for articulated deformation of humans without re-encoding the deformed shapes.
In addition to the generalization and editing capabilities, our representation includes all advantages of implicit surface modeling. 
Our patches can represent shapes of any arbitrary topology, and our reconstructions can be extracted at any arbitrary resolution using \emph{Marching Cubes}~\cite{lorensen1987marching}. 
Similar to DeepSDF~\cite{park2019deepsdf}, our network uses an auto-decoder architecture, combining classical optimization with learning, resulting in high-quality geometry.

\section{Related Work}

Our patch-based representation relates to many existing data structures and approaches in classical and learning-based visual computing. 
In the following, we focus on the most relevant existing representations, methods and applications.

\paragraph{Global Data Structures.} %
There are multiple widely-used data structures for geometric deep learning such as voxel grids \cite{Choy2016}, point clouds \cite{qi2017pointnet++}, meshes \cite{Wang2018} and implicit functions \cite{park2019deepsdf}.
To alleviate the memory limitations and speed-up training, improved versions of voxel grids with hierarchical space partitioning \cite{riegler2017octnet} and tri-linear interpolation \cite{ShimadaDispVoxNets2019} were recently proposed.
A mesh is an explicit discrete surface representation which can be useful in monocular rigid 3D reconstruction \cite{Wang2018,Kato2018}. 
Combined with patched-based policies, this representation can suffer from stitching artefacts \cite{Groueix2018}. 
All these data structures enable a limited level of detail given a constant memory size. 
In contrast, other representations such as sign distance functions (SDF) \cite{CurlessLevoy1996} represent surfaces implicitly as the zero-crossing of a volumetric level set function.

Recently, neural counterparts of implicit representations and approaches operating on them were proposed in the literature  \cite{park2019deepsdf,mescheder2019occupancy,chen2019learning,michalkiewicz2019deep}. 
Similarly to SDFs, these methods extract surfaces as zero level sets or decision boundaries, while differing in the type of the learned function. 
Thus, DeepSDF is a learnable variant of SDFs \cite{park2019deepsdf}, whereas Mescheder \textit{et  al.}~\cite{mescheder2019occupancy} train a spatial classifier (indicator function) for regions inside and outside of the scene. 
In theory, both methods allow for surface extraction at unlimited resolution.
Neural implicit functions have already demonstrated their effectiveness and robustness in many follow-up works and applications such as single-view 3D reconstruction  \cite{saito2019pifu,liu2019learning} as well as static \cite{sitzmann2019scene} and dynamic  \cite{niemeyer2019occupancy} object representation. 
While SAL \cite{Atzmon_2020_CVPR} perform shape completion from noisy full raw scans, one of our applications is shape completion from partial data with local refinement.
Unlike the aforementioned global approaches, PatchNets generalize much better, for example to new categories.

\paragraph{Patch-Based Representations.} 
Ohtake \textit{et al.}~\cite{ohtake2003multi} use a combination of implicit functions for  versatile shape representation and editing. 
Several neural techniques use mixtures of geometric primitives as well  \cite{tulsiani2017learning,genova2019learning,deng2019cvxnets,deprelle2019learning,williams2019voronoinet}. 
The latter have been shown as helpful abstractions in such tasks as shape segmentation, interpolation, classification and recognition, as well as 3D reconstruction. 
Tulsiani \textit{et al.}~\cite{tulsiani2017learning} learn to assemble shapes of various categories from explicit 3D geometric primitives (\textit{e.g.,} cubes and cuboids). 
Their method discovers a consistent structure and allows to establish semantic correspondences between the samples. 
Genova \textit{et al.}~\cite{genova2019learning} further develop the idea and learn a general template from data which is composed of implicit functions with local support. 
Due to the function choice, \textit{i.e.,} scaled axis-aligned anisotropic 3D Gaussians, shapes with sharp edges and thin structures are challenging for their method. 
In \textit{CVXNets} \cite{deng2019cvxnets}, solid objects are assembled in a piecewise manner from convex elements. 
This results in a differentiable form which is directly usable in physics and graphics engines. 
Deprelle \textit{et al.}~\cite{deprelle2019learning} decompose shapes into learnable combinations of deformable elementary 3D structures. 
VoronoiNet \cite{williams2019voronoinet} is a deep generative network which operates on a differentiable version of Voronoi diagrams. %
The concurrent NASA method \cite{deng2019nasa} focuses on articulated deformations, which is one of our applications.
In contrast to other patch-based approaches, our learned patches are not limited to hand-crafted priors but instead are more flexible and expressive.

\section{Proposed Approach} 

We represent the surface of any object as a combination of several surface patches. 
The patches form a mid-level representation, where each patch  represents the surface within a specified radius from its center.
This representation is generalizable across object categories, as most objects share similar geometry at the patch level. 
In the following, we explain how the patches are represented using artificial neural networks, the losses required to train such networks, as well as the algorithm to combine multiple patches for smooth surface reconstruction. 
\subsection{Implicit Patch Representation}
We represent a full object $i$ as a collection of $N_P=30$ patches. 
A patch $p$ represents a surface within a sphere of radius $r_{i,p} \in \mathbb{R}$, centered at $\mathbf{c}_{i,p} \in \mathbb{R}^3$.
Each patch can be oriented by a rotation about a canonical frame, parametrized by Euler angles $\mathbf{\phi}_{i,p}\in \mathbb{R}^3$. 
Let $\mathbf{e}_{i,p} = (r_{i,p}, \mathbf{c}_{i,p}, \mathbf{\phi}_{i,p}) \in \mathbb{R}^7$ denote all extrinsic patch parameters. 
Representing the patch surface in a canonical frame of reference lets us normalize the query 3D point, leading to more object-agnostic and generalizable patches. 

The patch surface is represented as an implicit signed-distance function (SDF), which maps 3D points to their signed distance from the closest surface. 
This offers several advantages, as these functions are a continuous representation of the surface, unlike point clouds or meshes.
In addition, the surface can be extracted at any resolution without large memory requirement, unlike for voxel grids. 
In contrast to prior work \cite{williams2019voronoinet,genova2019learning}, which uses simple patch primitives, we parametrize the patch surface as a neural network (PatchNet).
Our network architecture is based on the auto-decoder of DeepSDF~\cite{park2019deepsdf}. 
The input to the network is a \emph{patch latent code} $\mathbf{z}\in \mathbb{R}^{N_z}$ of length $N_z=128$, which describes the patch surface, and a 3D query point $\mathbf{x} \in \mathbb{R}^3$.
The output is the scalar SDF value of the surface at $\mathbf{x}$.
Similar to DeepSDF, we use eight weight-normalized~\cite{salimans2016weight} fully-connected layers with 128 output dimensions and ReLU activations, and we also concatenate $\mathbf{z}$ and $\mathbf{x}$ to the input of the fifth layer. The last fully-connected layer outputs a single scalar to which we apply \emph{tanh} to obtain the SDF value.

\subsection{Preliminaries}\label{sec:prelim}

\paragraph{Preprocessing:}
Given a watertight mesh, we preprocess it to obtain SDF values for 3D point samples. 
First, we center each mesh and fit it tightly into the unit sphere.
We then sample points, mostly close to the surface, and compute their truncated signed distance to the object surface, with truncation at $0.1$. 
For more details on the sampling strategy, please refer to %
\cite{park2019deepsdf}.

\paragraph{Auto-Decoding:}
Unlike the usual setting, we do not use an encoder that regresses patch latent codes and extrinsics.
Instead, we follow DeepSDF~\cite{park2019deepsdf} and auto-decode shapes: we treat the patch latent codes and extrinsics of each object as free variables to be optimized for during training.
I.e., instead of back-propagating into an encoder, we employ the gradients to learn these parameters directly during training.

\paragraph{Initialization:}\label{sec:initialization}

Since we perform auto-decoding, we treat the patch latent codes and extrinsics as free variables, similar to classical optimization.
Therefore, we can directly initialize them.
All patch latent codes are initially set to zero, and the patch positions are initialized by greedy farthest point sampling of point samples of the object surface. 
We set each patch radius to the minimum such that each surface point sample is covered by its closest patch. 
The patch orientation aligns the $z$-axis of the patch coordinate system with the surface normal.

\subsection{Loss Functions} %

We train PatchNet by auto-decoding $N$ full objects. 
The patch latent codes of an object~$i$ are %
$\mathbf{z}_i = [\mathbf{z}_{i,0},\mathbf{z}_{i,1},\ldots,\mathbf{z}_{i,N_P-1}]$, with each patch latent code of length $N_z$. 
Patch extrinsics are represented as 
$\mathbf{e}_i = [\mathbf{e}_{i,0}, \mathbf{e}_{i,1}, \ldots, \mathbf{e}_{i,N_P-1}].$ 
Let $\theta$ denote the trainable weights of PatchNet. 
We employ the following loss function:
\begin{align}
    \mathcal{L}(\mathbf{z}_i, \mathbf{e}_i, \theta) = \mathcal{L}_\text{recon}(\mathbf{z}_i, \mathbf{e}_i, \theta) + \mathcal{L}_\text{ext}(\mathbf{e}_i) + 
     \mathcal{L}_\text{reg}(\mathbf{z}_i)
\enspace{.}
    \label{eq:loss_main}
\end{align}
Here, $\mathcal{L}_{recon}$ is the surface reconstruction loss, $\mathcal{L}_{ext}$ is the extrinsic loss guiding the extrinsics for each patch, and $\mathcal{L}_{reg}$ is a regularizer on the patch latent codes. 
\paragraph{Reconstruction Loss:}
The reconstruction loss minimizes the SDF values between the predictions and the ground truth for each patch: 
\begin{align}
    \mathcal{L}_\text{recon}(\mathbf{z}_i, \mathbf{e}_i, \theta) = \frac{1}{N_P} \sum_{p=0}^{N_P-1} \frac{1}{|S(\mathbf{e}_{i,p})|} \sum_{\mathbf{x} \in S(\mathbf{e}_{i,p})} \big\| f(\mathbf{x},\mathbf{z}_{i,p},\theta) - s(\mathbf{x}) \big\|_1, 
\end{align}
where $f(\cdot)$ and $s(\mathbf{x})$ denote a forward pass of the network and the ground truth truncated SDF values at point $\mathbf{x}$, respectively; 
$S(\mathbf{e}_{i,p})$ is the set of all (normalized) point samples that lie within the bounds of patch $p$ with extrinsics $\mathbf{e}_{i,p}$. 
\paragraph{Extrinsic Loss:}

The composite extrinsic loss ensures all patches contribute to the surface and are placed such that the surfaces are learned in a canonical space: 
\begin{align}
    \mathcal{L}_\text{ext}(\mathbf{e}_i) = \mathcal{L}_\text{sur}(\mathbf{e}_i) + \mathcal{L}_\text{cov}(\mathbf{e}_i) +
    \mathcal{L}_\text{rot}(\mathbf{e}_i) +
    \mathcal{L}_\text{scl}(\mathbf{e}_i) + 
    \mathcal{L}_\text{var}(\mathbf{e}_i)
    \enspace{.}
    \label{eq:loss}
\end{align}
$\mathcal{L}_\text{sur}$ ensures that every patch stays close to the surface: 
\begin{align}
    \mathcal{L}_\text{sur}(\mathbf{e}_i) = \omega_\text{sur}\cdot\frac{1}{N_P} \sum_{p=0}^{N_P-1} \max (\min_{\mathbf{x} \in \mathbf{O}_{i}} \big\| \mathbf{c}_{i,p} - \mathbf{x} \big\|_2^2, t)
    \enspace{.}
\end{align}
Here, $\mathbf{O}_{i}$ is the set of surface points of object $i$.
We use this term only when the distance between a patch and the surface is greater than a threshold $t=0.06$. 

A symmetric coverage loss $\mathcal{L}_\text{cov}$ encourages each point on the surface to be covered by a at least one patch: 
\begin{align}
    \mathcal{L}_\text{cov}(\mathbf{e}_i) =\omega_\text{cov}\cdot \frac{1}{|\mathbf{U}_i|} \sum_{\mathbf{x} \in \mathbf{U}_i} \frac{w_{i,p,\mathbf{x}}}{\sum_p w_{i,p,\mathbf{x}}} (\big\| \mathbf{c}_{i,p} - \mathbf{x} \big\|_2 - r_{i,p})
    \enspace{,}
\end{align}
where $\mathbf{U}_i \subseteq \mathbf{O}_{i}$ are all surface points that are not covered by any patch, \textit{i.e.}, outside the bounds of all patches. 
$w_{i,p,\mathbf{x}}$ weighs the patches based on their distance from $\mathbf{x}$, with $w_{i,p,\mathbf{x}} = \exp{(-0.5 \cdot  ((\big\| \mathbf{c}_{i,p} - \mathbf{x} \big\|_2 - r_{i,p})/\sigma)^2 )}$ where $\sigma=0.05$. 

We also introduce a loss to align the patches with the surface normals.
This encourages the patch surface to be learned in a canonical frame of reference: 
\begin{align}
    \mathcal{L}_\text{rot}(\mathbf{e}_i) = \omega_\text{rot}\cdot\frac{1}{N_P} \sum_{p=0}^{N_P-1} (1 - \langle\mathbf{\phi}_{i,p} \cdot [0,0,1]^T, \mathbf{n}_{i,p}\rangle)^2
    \enspace{.}
\end{align}
Here, $\mathbf{n}_{i,p}$ is the surface normal at the point $\mathbf{o}_{i,p}$ closest to the patch center, \textit{i.e.,} $\mathbf{o}_{i,p} =  \argmin\limits_{{\mathbf{x} \in \mathbf{O}_i}} \big\| \mathbf{x} - \mathbf{c}_{i,p} \big\|_2$.  

Finally, we introduce two losses for the extent of the patches. The first loss encourages the patches to be reasonably small. 
This prevents significant overlap between different patches: 
\begin{align}
    \mathcal{L}_\text{scl}(\mathbf{e}_i) = \omega_\text{scl}\cdot\frac{1}{N_P} \sum_{p=0}^{N_P-1} r_{i,p}^2
    \enspace{.}
\end{align}
The second loss encourages all patches to be of similar sizes. 
This prevents the surface to be reconstructed only using very few large patches: 
\begin{align}
    \mathcal{L}_\text{var}(\mathbf{e}_i) = \omega_\text{var}\cdot\frac{1}{N_P} \sum_{p=0}^{N_P-1} (r_{i,p} - m_i)^2
    \enspace{,}
\end{align}
where $m_i$ is the mean patch radius of object $i$.

\paragraph{Regularizer:} 
Similar to DeepSDF, we add an $\ell_2$-regularizer on the latent codes assuming a Gaussian prior distribution: 
\begin{align}
    \mathcal{L}_\text{reg}(\mathbf{z}_i) =  \omega_\text{reg}\cdot\frac{1}{N_P} \sum_{p=0}^{N_P-1} \big\| \mathbf{z}_{i,p} \big\| _2^2
    \enspace{.}
\end{align}

\paragraph{Optimization:}
At training time, we optimize the following problem: 
\begin{align}
    \argmin_{\theta, \{\mathbf{z}_i\}_i, \{\mathbf{e}_i\}_i} \sum_{i=0}^{N-1} \mathcal{L}(\mathbf{z}_i, \mathbf{e}_i, \theta) 
     \enspace{.}
    \label{eq:opt}
\end{align}

At test time, we can reconstruct any surface using our learned patch-based representation.
Using the same initialization of extrinsics and patch latent codes, and given point samples with their SDF values, we optimize for the patch latent codes and the patches extrinsics with fixed network weights.

\subsection{Blended Surface Reconstruction}

For a smooth surface reconstruction of object $i$, \emph{e.g.} for Marching Cubes, we blend between different patches in the overlapping regions to obtain the blended SDF prediction $g_i(\mathbf{x})$. Specifically, $g_i(\mathbf{x})$ is computed as a weighted linear combination of the SDF values $f(\mathbf{x},\mathbf{z}_{i,p},\theta)$ of the overlapping patches: 
\begin{align}\label{eq:mixture}
    g_i(\mathbf{x}) = \sum_{p\in P_{i,\mathbf{x}}} \frac{w_{i,p,\mathbf{x}}}{\sum_{p\in P_{i,\mathbf{x}}} w_{i,p,\mathbf{x}}} f(\mathbf{x},\mathbf{z}_{i,p},\theta), 
\end{align}
with $P_{i,\mathbf{x}}$ denoting the patches which overlap at point $\mathbf{x}$.
For empty $P_{i,\mathbf{x}}$, we set $g_i(\mathbf{x})=1$.
The blending weights are defined as:
\begin{align}
    w_{i,p, \mathbf{x}} = \operatorname{exp} \bigg(-\frac{1}{2}  \bigg(\frac{\big\| \mathbf{c}_{i,p} - \mathbf{x} \big\|_2}{\sigma}\bigg)^2 \bigg) - \operatorname{exp} \bigg(-\frac{1}{2} \bigg(\frac{r_{i,p}}{\sigma}\bigg)^2 \bigg),
\end{align}
with $\sigma = r_{i,p}/3$. 
The offset ensures that the weight is zero at the patch boundary.

\section{Experiments} 
In the following, we show the effectiveness of our patch-based representation on several different problems. For an ablation study of the loss functions, please refer to the supplemental.

\subsection{Settings}

\subsubsection{Datasets}
We employ \emph{ShapeNet}~\cite{chang2015shapenet} for most experiments.
We perform preprocessing with the code of Stutz~\etal~\cite{stutz2018learning}, similar to \cite{mescheder2019occupancy,genova2019deep}, 
to make the meshes watertight and normalize them within a unit cube. 
For training and test splits, we follow Choy~\etal~\cite{Choy2016}.
The results in Tables \ref{tab:surfaceRecon} and \ref{tab:surfaceReconComparison} use the full test set. 
Other results refer to a reduced test set, where we randomly pick 50 objects from each of the 13 categories.
In the supplemental, we show that our results on the reduced test set are representative of the full test set.
In addition, we use Dynamic FAUST~\cite{Bogo2017} for testing. We subsample the test set from DEMEA~\cite{Tretschk2019arXiv} by concatenating all test sequences and taking every 20th mesh.
We generate 200k SDF point samples per shape during preprocessing.

\subsubsection{Metrics}
We use three error metrics. For Intersection-over-Union (IoU), higher is better. For Chamfer distance (Chamfer), lower is better. For F-score, higher is better. The supplementary material contains further details on these metrics.

\subsubsection{Training Details}

We train our networks using \emph{PyTorch}~\cite{pytorch}. 
The number of epochs is 1000, the learning rate for the network is initially $5\cdot10^{-4}$, and for the patch latent codes and extrinsics $10^{-3}$. We half both learning rates every 200 epochs.
For optimization, we use Adam~\cite{adam} and a batch size of 64. 
For each object in the batch, we randomly sample 3k SDF point samples.
The weights for the losses are: $\omega_\text{scl}=0.01$, $\omega_\text{var}=0.01$, $\omega_\text{sur}=5$, $\omega_\text{rot}=1$, $\omega_\text{sur}=200$. 
We linearly increase $\omega_\text{reg}$ from 0 to $10^{-4}$ for 400 epochs and then keep it constant.

\subsubsection{Baseline}

We design a ``global-patch" baseline similar to DeepSDF, which only uses a single patch without extrinsics. 
The patch latent size is 4050, matching ours.
The learning rate scheme is the same as for our method.  

\subsection{Surface Reconstruction}

We first consider surface reconstruction.
\subsubsection{Results}
\begin{figure}
\centering
\includegraphics[trim={340 0 340 0} ,clip,height=2.4cm]
{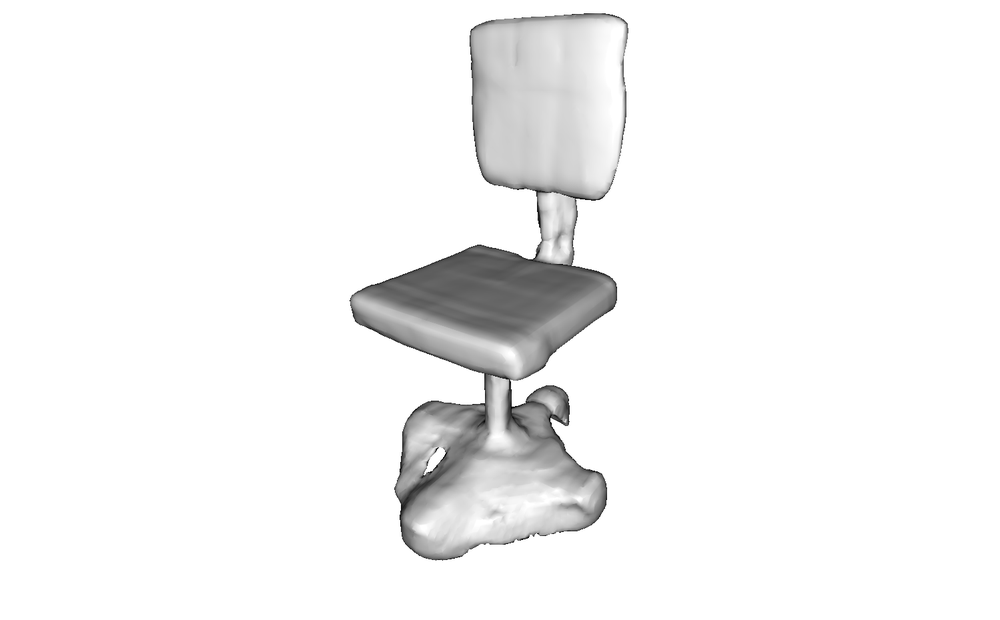}
\includegraphics[trim={340 0 340 0} ,clip,height=2.4cm]
{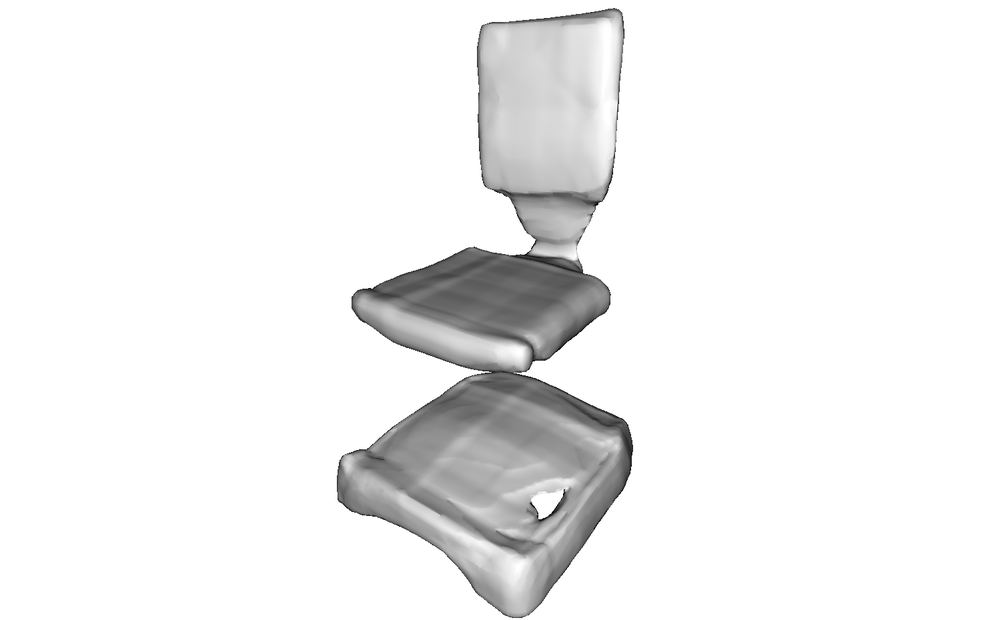}
\includegraphics[trim={340 0 340 0} ,clip,height=2.4cm]
{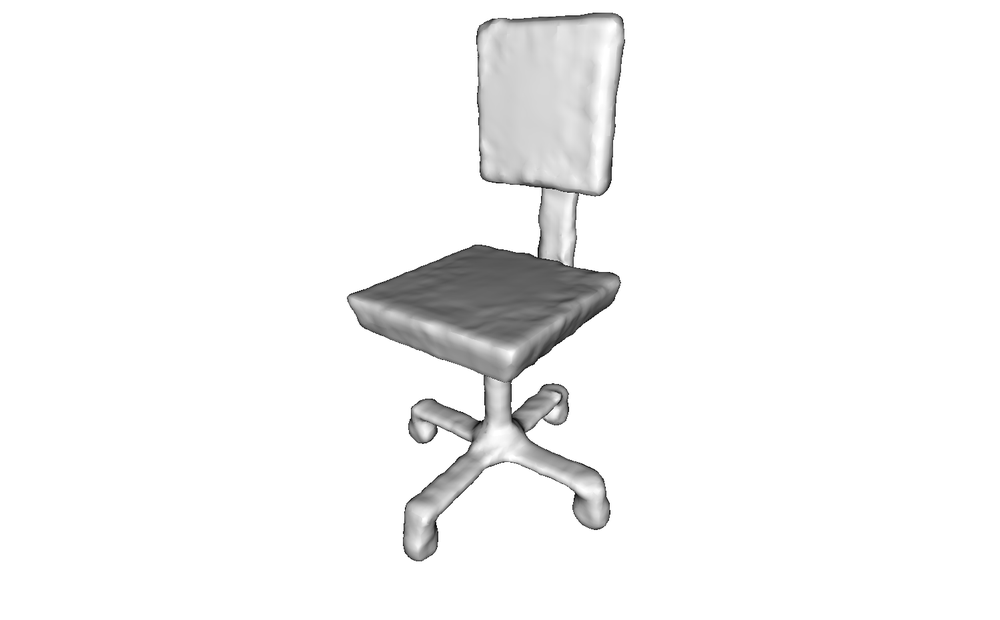}
\includegraphics[trim={340 40 200 30} ,clip,height=2.4cm]
{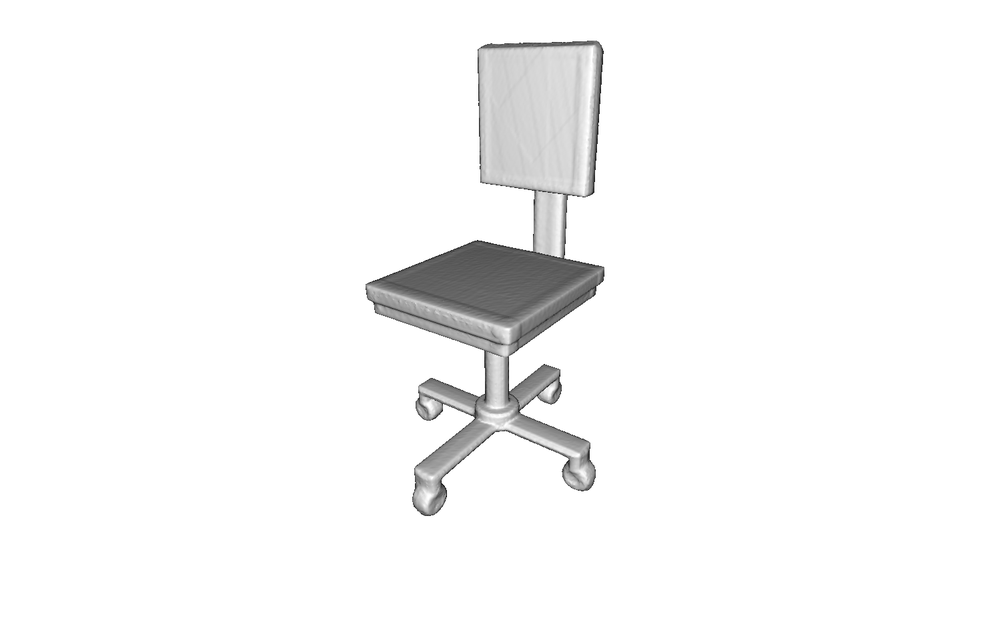}
\includegraphics[trim={300 -170 240 0} ,clip,height=2.4cm]
{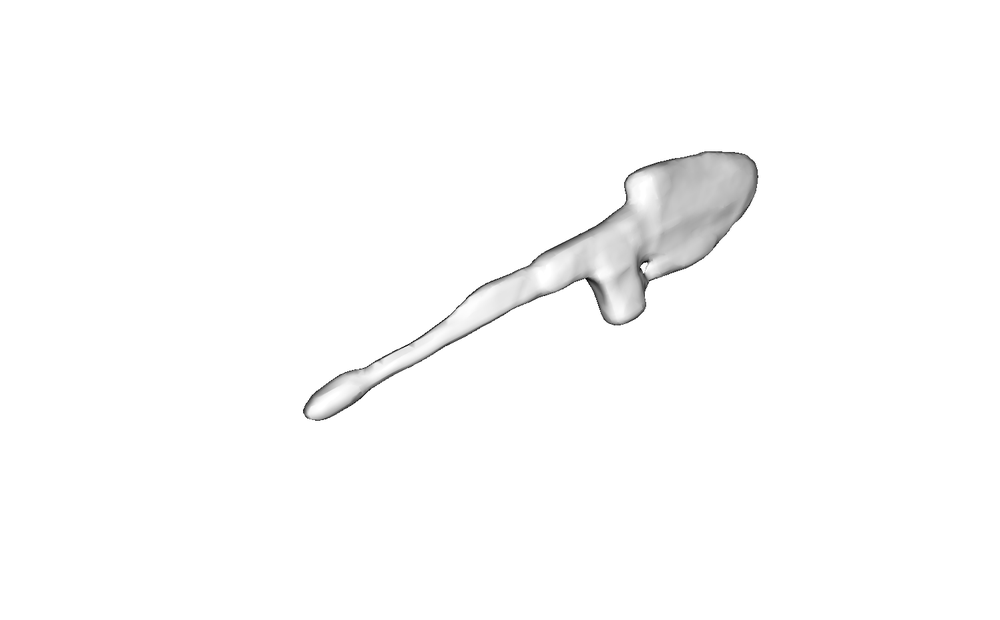}
\includegraphics[trim={300 -170 240 0} ,clip,height=2.4cm]
{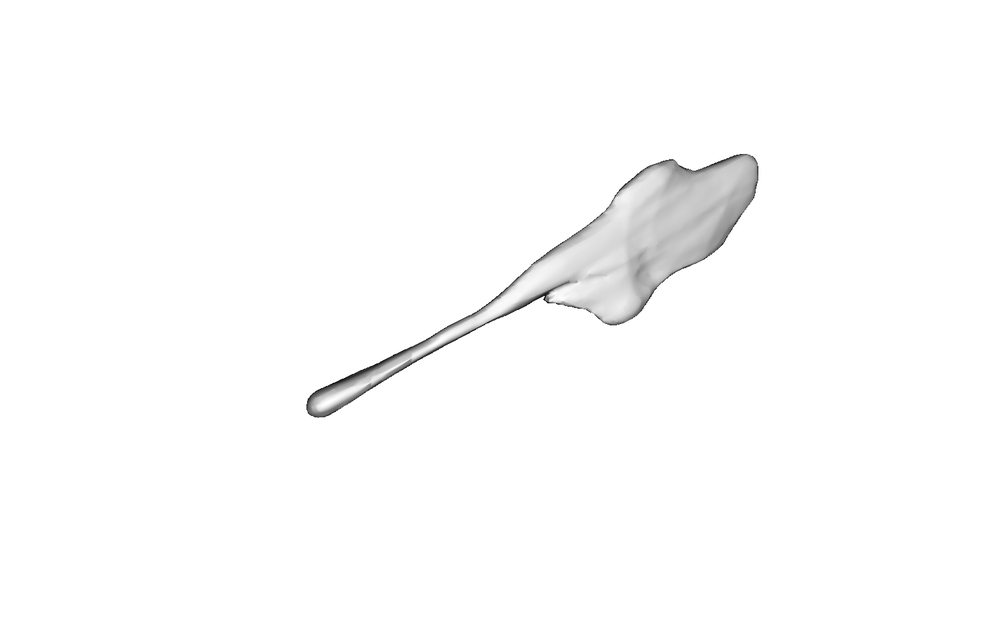}
\includegraphics[trim={300 -170 240 0} ,clip,height=2.4cm]
{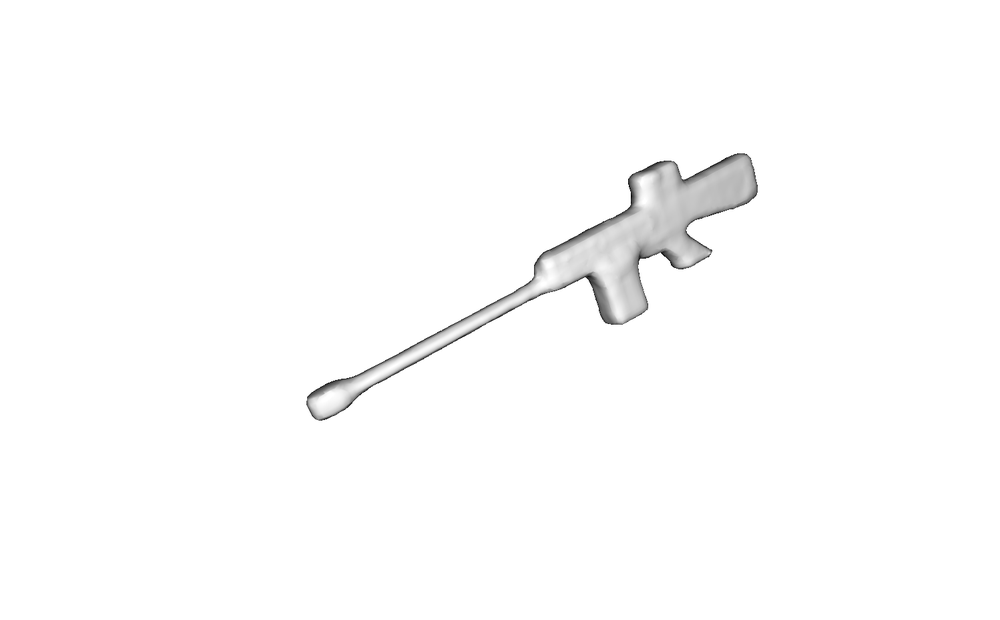}
\includegraphics[trim={300 -190 90 -60} ,clip,height=2.4cm]
{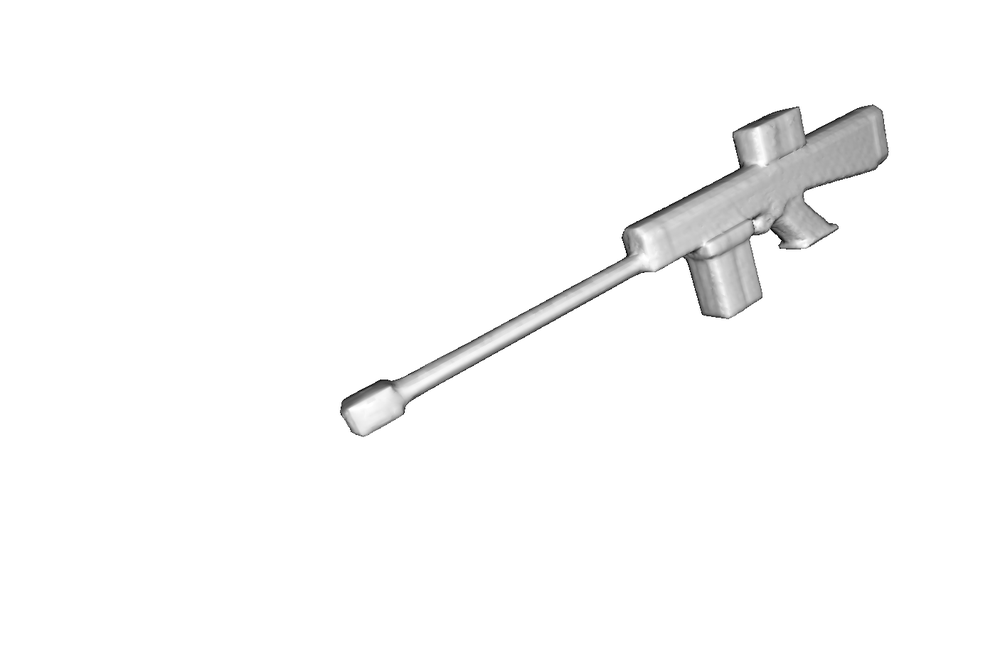}

\caption{Surface Reconstruction. From left to right: DeepSDF, baseline, ours, groundtruth.}
\label{fig:qualitative}
\end{figure}

We train our approach on a subset of the training data, where we randomly pick $100$ shapes from each category. 
In addition to comparing with our baseline, we compare with DeepSDF~\cite{park2019deepsdf} as setup in their paper. 
Both DeepSDF and our baseline use the subset.
Qualitative results are shown in Fig.~\ref{fig:qualitative} and \ref{fig:generalization}.

\begin{table}
\centering
\caption{Surface Reconstruction. We significantly outperform DeepSDF~\cite{park2019deepsdf} and our baseline on all categories of ShapeNet almost everywhere.}
\setlength\tabcolsep{3pt} %
\resizebox{0.8\columnwidth}{!}{
\begin{tabular}{l|ccc|ccc|ccc} 
\hline
 & \multicolumn{3}{c}{IoU}  & \multicolumn{3}{c}{Chamfer} & \multicolumn{3}{c}{F-score} \\
                          Category &  DeepSDF & Baseline & Ours      &  DeepSDF & Baseline & Ours  & DeepSDF & Baseline & Ours \\
\hline
airplane                  &  84.9 & 65.3 & \textbf{91.1} &         0.012 & 0.077 & \textbf{0.004} &   83.0 & 72.9 & \textbf{97.8}\\
bench                     &  78.3 & 68.0 & \textbf{85.4} &         0.021 & 0.065 & \textbf{0.006} &   91.2 & 80.6 & \textbf{95.7}\\
cabinet                   &  92.2 & 88.8 & \textbf{92.9} &         \textbf{0.033} & 0.055 & 0.110 &   \textbf{91.6} & 86.4 & 91.2\\
car                       &  87.9 & 83.6 & \textbf{91.7} &         \textbf{0.049} & 0.070 & \textbf{0.049} &   82.2 & 74.5 & \textbf{87.7}\\
chair                     &  81.8 & 72.9 & \textbf{90.0} &         0.042 & 0.110 & \textbf{0.018} &   86.6 & 75.5 & \textbf{94.3}\\
display                   &  91.6 & 86.5 & \textbf{95.2} &         \textbf{0.030} & 0.061 & 0.039 &   93.7 & 87.0 & \textbf{97.0}\\
lamp                      &  74.9 & 63.0 & \textbf{89.6} &         0.566 & 0.438 & \textbf{0.055} &   82.5 & 69.4 & \textbf{94.9}\\
rifle                     &  79.0 & 68.5 & \textbf{93.3} &         0.013 & 0.039 & \textbf{0.002} &   90.9 & 82.3 & \textbf{99.3}\\
sofa                      &  92.5 & 85.4 & \textbf{95.0} &         0.054 & 0.226 & \textbf{0.014} &   92.1 & 84.2 & \textbf{95.3}\\
speaker                   &  91.9 & 86.7 & \textbf{92.7} &         \textbf{0.050} & 0.094 & 0.243 &   87.6 & 79.4 & \textbf{88.5}\\
table                     &  84.2 & 71.9 & \textbf{89.4} &         0.074 & 0.156 & \textbf{0.018} &   91.1 & 79.2 & \textbf{95.0}\\
telephone                 &  96.2 & 95.0 & \textbf{98.1} &         0.008 & 0.016 & \textbf{0.003} &   97.7 & 96.2 & \textbf{99.4}\\
watercraft                &  85.2 & 79.1 & \textbf{93.2} &         0.026 & 0.041 & \textbf{0.009} &   87.8 & 80.2 & \textbf{96.4}\\
\hline
mean                      &  77.4 & 76.5 & \textbf{92.1} &         0.075 & 0.111 & \textbf{0.044} &   89.9  & 80.6 & \textbf{94.8}  \\
\hline
\end{tabular}
\label{tab:surfaceRecon}} %
\end{table}

Table~\ref{tab:surfaceRecon} shows the quantitative results for surface reconstruction.
We significantly outperform DeepSDF and our baseline almost everywhere, demonstrating the higher-quality afforded by our patch-based representation. 

We also compare with several state-of-the-art approaches on implicit surface reconstruction, OccupancyNetworks~\cite{mescheder2019occupancy}, Structured Implicit Functions~\cite{genova2019learning} and Deep Structured Implicit Functions~\cite{genova2019deep}\footnote{DSIF is also known as \emph{Local Deep Implicit Functions for 3D Shape}.}.
While they are trained on the full ShapeNet shapes, we train our model only on a small subset.
Even in this disadvantageous and challenging setting, we outperform these approaches on most categories, see Table~\ref{tab:surfaceReconComparison}.
Note that we compute the metrics consistently with Genova \etal~\cite{genova2019deep} and thus can directly compare to numbers reported in their paper.
\begin{table}
\centering
\caption{Surface Reconstruction. We outperform OccupancyNetworks (OccNet)~\cite{mescheder2019occupancy}, Structured Implicit Functions (SIF)~\cite{genova2019learning}, and Deep Structured Implicit Functions (DSIF)~\cite{genova2019deep} almost everywhere.}
\setlength\tabcolsep{3pt} %
\resizebox{0.8\columnwidth}{!}{
\begin{tabular}{l|cccc|cccc|cccc} 
\hline
 & \multicolumn{4}{c}{IoU}  & \multicolumn{4}{c}{Chamfer} & \multicolumn{4}{c}{F-score} \\
                          Category & OccNet & SIF  &   DSIF  & Ours      & OccNet          & SIF    &           DSIF &  Ours & OccNet & SIF  & DSIF & Ours \\
\hline
airplane                  &  77.0  & 66.2 & \textbf{91.2} & 91.1 &         0.016   &  0.044 & 0.010 & \textbf{0.004} &  87.8  & 71.4 & 96.9 & \textbf{97.8}\\
bench                     &  71.3  & 53.3 & \textbf{85.6} & 85.4 &         0.024   &  0.082 & 0.017 & \textbf{0.006} &  87.5  & 58.4 & 94.8 & \textbf{95.7}\\
cabinet                   &  86.2  & 78.3 & \textbf{93.2} & 92.9 &         0.041   &  0.110 & \textbf{0.033} & 0.110 &  86.0  & 59.3 & \textbf{92.0} & 91.2\\
car                       &  83.9  & 77.2 & 90.2 & \textbf{91.7} &         0.061   &  0.108 & \textbf{0.028} & 0.049 &  77.5  & 56.6 & 87.2 & \textbf{87.7}\\
chair                     &  73.9  & 57.2 & 87.5 & \textbf{90.0} &         0.044   &  0.154 & 0.034 & \textbf{0.018} &  77.2  & 42.4 & 90.9 & \textbf{94.3}\\
display                   &  81.8  & 69.3 & 94.2 & \textbf{95.2} &         0.034   &  0.097 & \textbf{0.028} & 0.039 &  82.1  & 56.3 & 94.8 & \textbf{97.0}\\
lamp                      &  56.5  & 41.7 & 77.9 & \textbf{89.6} &         0.167   &  0.342 & 0.180 & \textbf{0.055} &  62.7  & 35.0 & 83.5 & \textbf{94.9} \\
rifle                     &  69.5  & 60.4 & 89.9 & \textbf{93.3} &         0.019   &  0.042 & 0.009 & \textbf{0.002} &  86.2  & 70.0 & 97.3 & \textbf{99.3}\\
sofa                      &  87.2  & 76.0 & 94.1 & \textbf{95.0} &         0.030   &  0.080 & 0.035 & \textbf{0.014} &  85.9  & 55.2 & 92.8 & \textbf{95.3}\\
speaker                   &  82.4  & 74.2 & 90.3 & \textbf{92.7} &         0.101   &  0.199 & \textbf{0.068} & 0.243 &  74.7  & 47.4 & 84.3 & \textbf{88.5}\\
table                     &  75.6  & 57.2 & 88.2 & \textbf{89.4} &         0.044   &  0.157 & 0.056 & \textbf{0.018} &  84.9  & 55.7 & 92.4 & \textbf{95.0} \\
telephone                 &  90.9  & 83.1 & 97.6 & \textbf{98.1} &         0.013   &  0.039 & 0.008 & \textbf{0.003} &  94.8  & 81.8 & 98.1 & \textbf{99.4}\\
watercraft                &  74.7  & 64.3 & 90.1 & \textbf{93.2} &         0.041   &  0.078 & 0.020 & \textbf{0.009} &  77.3  & 54.2 & 93.2 & \textbf{96.4} \\
\hline
mean                      &  77.8  & 66.0 & 90.0 & \textbf{92.1}  &         0.049   &  0.118 & \textbf{0.040} & 0.044 &  81.9  & 59.0 & 92.2 & \textbf{94.8}  \\
\hline
\end{tabular}
\label{tab:surfaceReconComparison}
} %
\end{table}

\subsubsection{Generalization}\label{sec:generalization}
\begin{figure}
\centering
\includegraphics[trim={400 0 360 0} ,clip,height=2.8cm]
{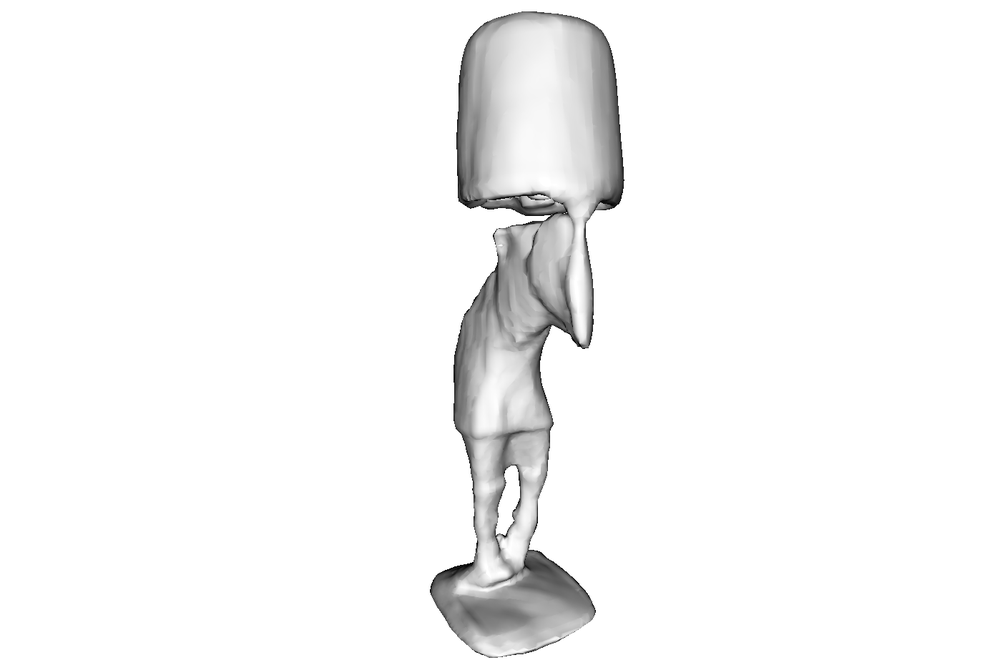}
\includegraphics[trim={400 0 360 0} ,clip,height=2.8cm]
{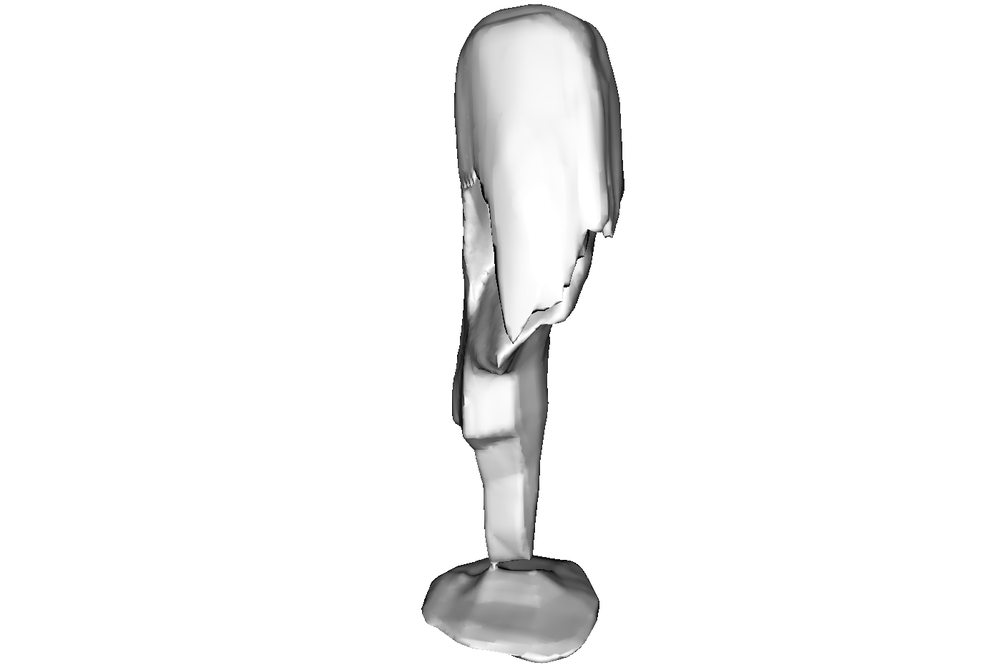}
\includegraphics[trim={0 0 0 0},clip,height=2.8cm]
{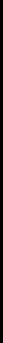}
\includegraphics[trim={400 0 360 0} ,clip,height=2.8cm]
{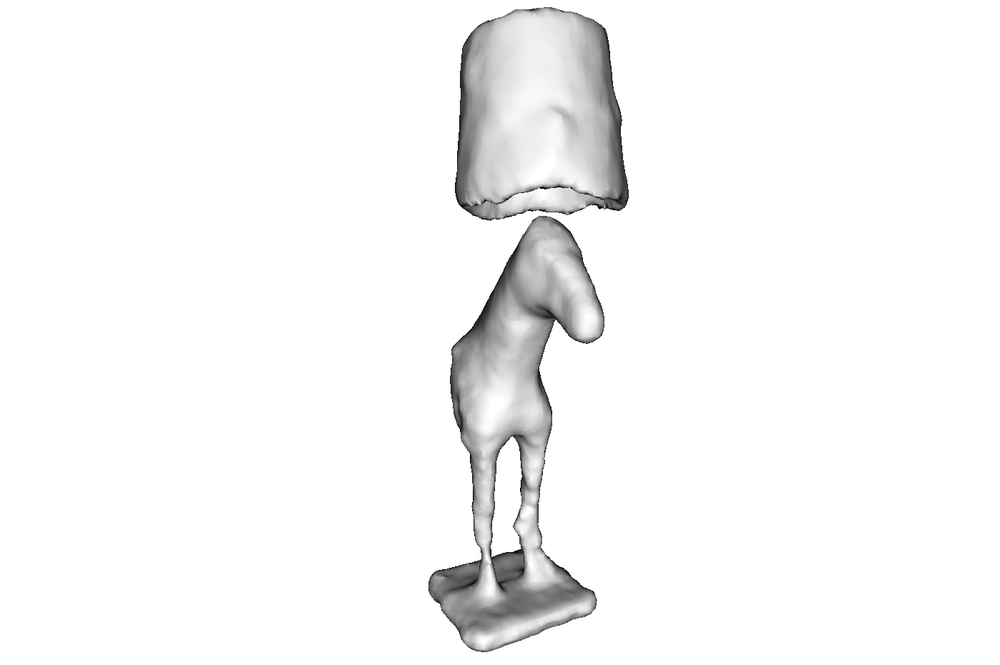}
\includegraphics[trim={400 0 360 0} ,clip,height=2.8cm]
{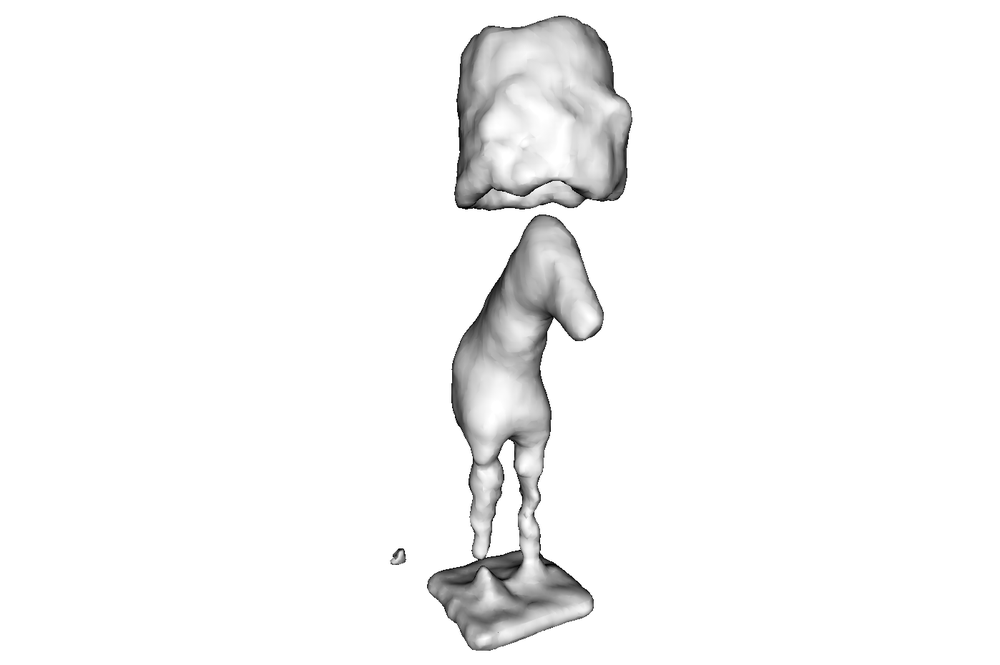}
\includegraphics[trim={0 0 0 0},clip,height=2.8cm]
{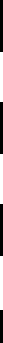}
\includegraphics[trim={400 0 360 0} ,clip,height=2.8cm]
{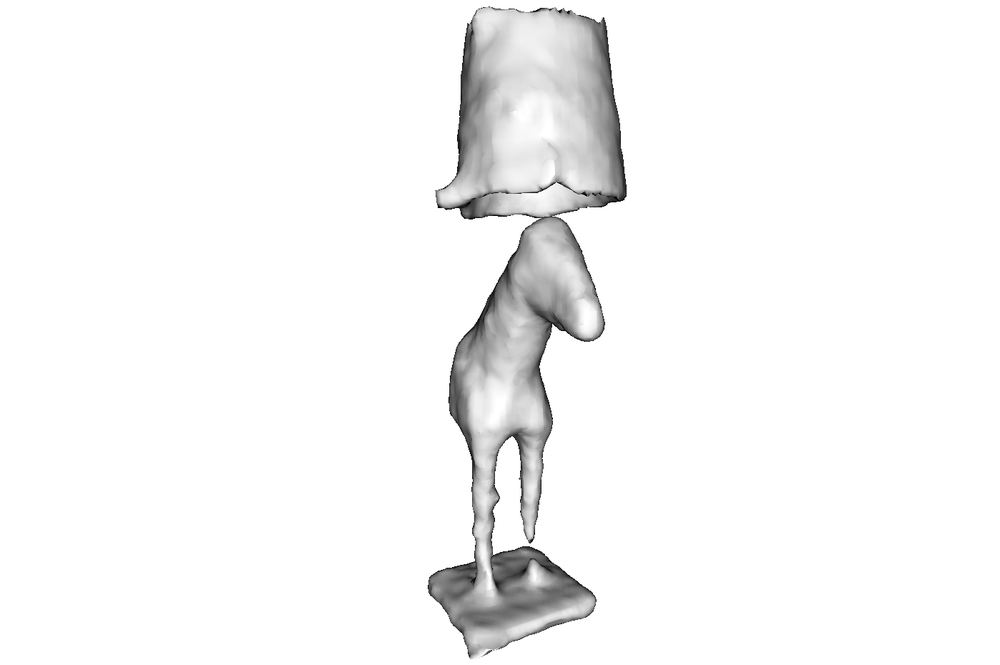}
\includegraphics[trim={400 0 360 0} ,clip,height=2.8cm]
{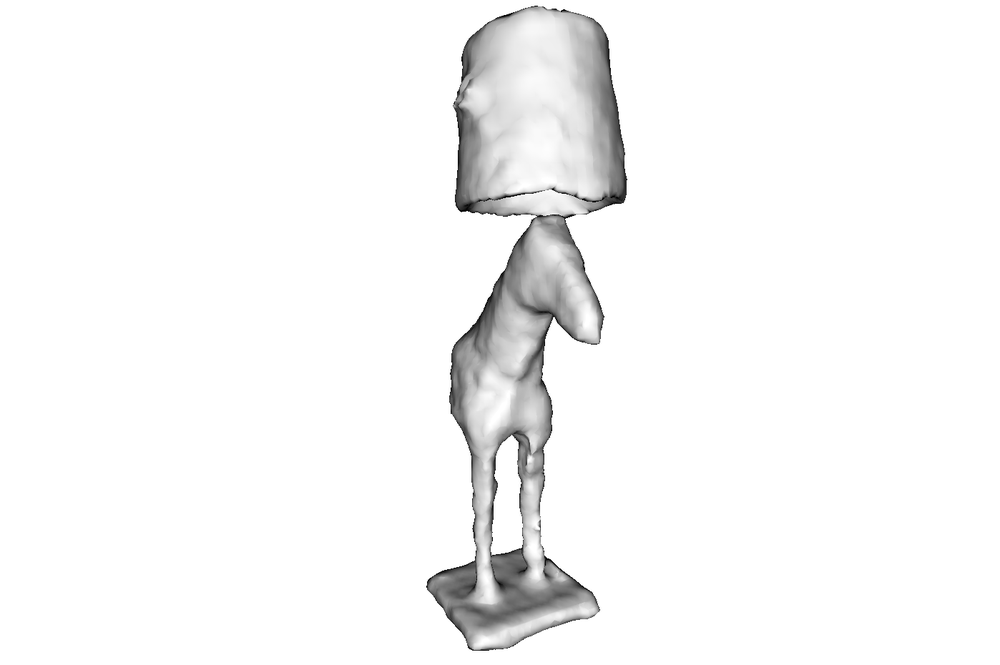}
\includegraphics[trim={400 0 360 0} ,clip,height=2.8cm]
{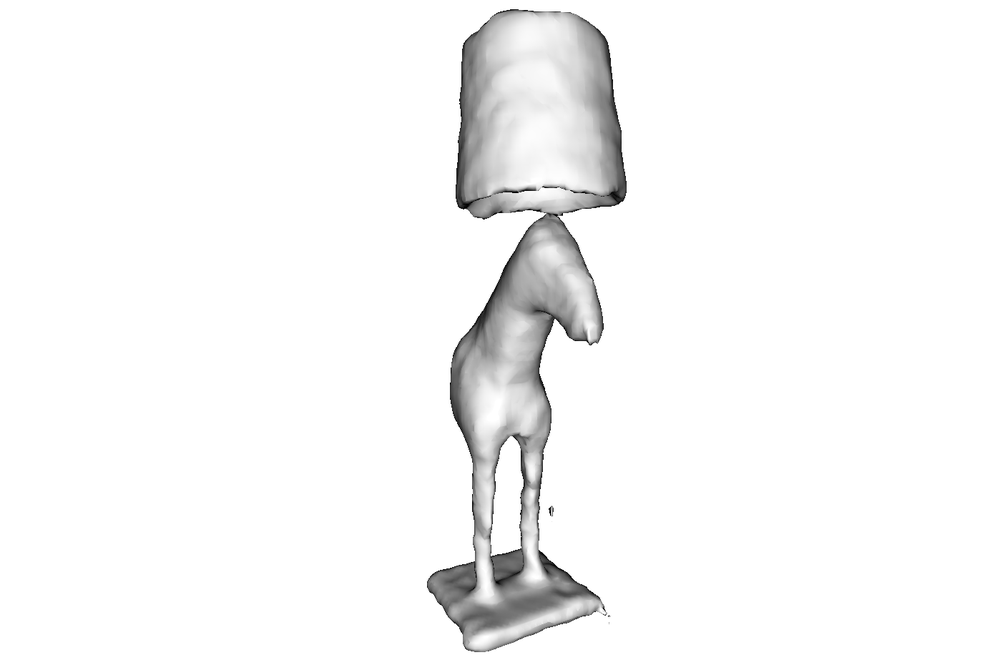}
\includegraphics[trim={400 0 360 0} ,clip,height=2.8cm]
{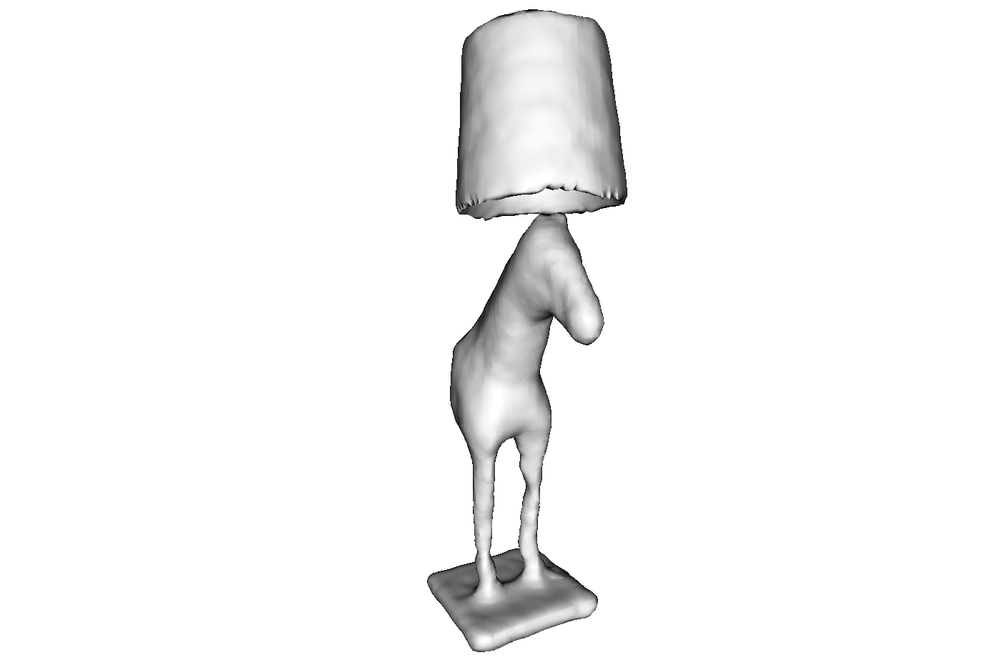}
\includegraphics[trim={400 0 360 0} ,clip,height=2.8cm]
{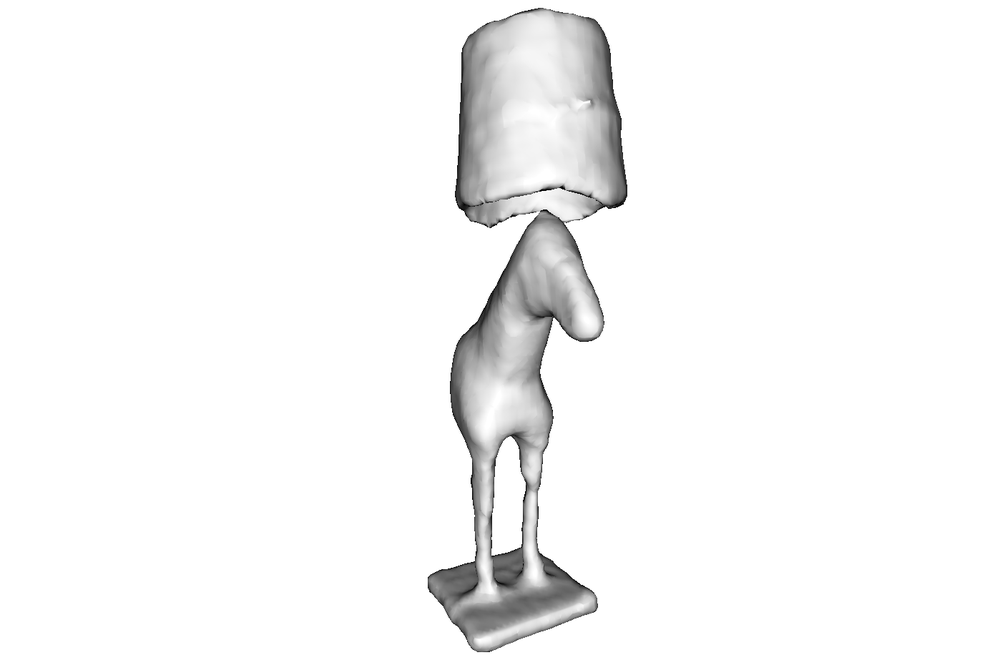}
\includegraphics[trim={0 0 0 0} ,clip,height=2.8cm]
{eccv2020kit/figures/black_bar.png}
\includegraphics[trim={400 -10 360 0} ,clip,height=2.8cm]
{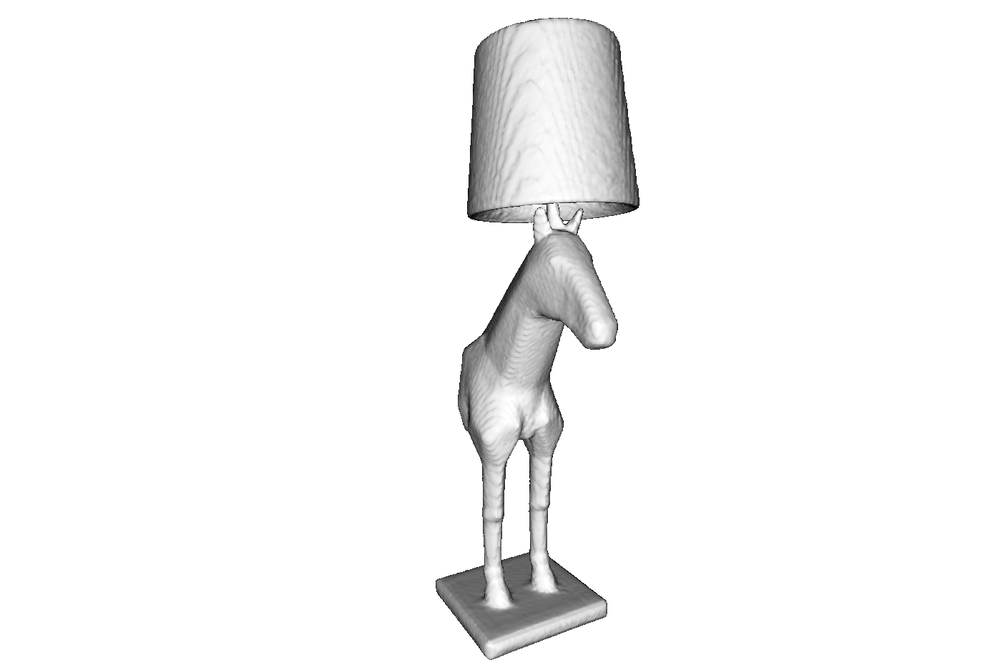}

\caption{Generalization. From left to right: DeepSDF, baseline, ours on one category, ours on one shape, ours on 1 shape per category, ours on 3 per category, ours on 10 per category, ours on 30 per category, ours on 100 per category, and groundtruth.}
\label{fig:generalization}
\end{figure}

Our patch-based representation is more generalizable compared to existing representations. 
To demonstrate this, we design several experiments with different training data.
We modify the learning rate schemes to equalize the number of network weight updates.
For each experiment, we compare our method with the baseline approaches described above. 
We use a reduced ShapeNet test set, which consists of $50$ shapes from each category.
Fig.~\ref{fig:generalization} shows qualitative results and comparisons.
We also show cross-dataset generalization by evaluating on $647$ meshes from the Dynamic FAUST~\cite{Bogo2017} test set.
In the first experiment, we train the network on shapes from the \emph{Cabinet} category and try to reconstruct shapes from every other category. 
We significantly outperform the baselines almost everywhere, see Table~\ref{tab:generalizationCategory}. 
The improvement is even more noticeable for cross dataset generalization with around $70\%$ improvement in the F-score compared to our global-patch baseline.
 
\begin{figure}
    \centering
    \includegraphics[width=\textwidth]{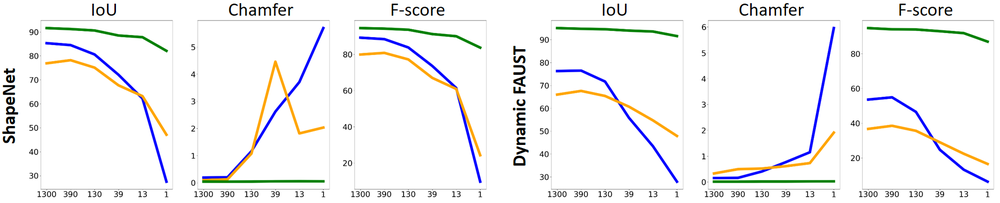}
    \caption{Generalization. We train our PatchNet (green), the global-patch baseline (orange), and DeepSDF (blue) on different numbers of shapes (x-axis). Results on different metrics on our reduced test sets are shown on the y-axis. For IoU and F-score, higher is better. For Chamfer distance, lower is better. %
    }
    \label{fig:gen-plots}
\end{figure}

In the second experiment, we evaluate the amount of training data required to train our network. 
We train both our network as well as the baselines on $30$, $10$, $3$ and $1$ shapes per-category of ShapeNet. 
In addition, we also include an experiment training the networks on a single randomly picked shape from ShapeNet.
Fig.~\ref{fig:gen-plots} shows the errors for ShapeNet (mean across categories) and Dynamic FAUST. 
The performance of our approach degrades only slightly with a decreasing number of training shapes.
However, the baseline approach of DeepSDF degrades much more severely.
This is even more evident for cross dataset generalization on Dynamic FAUST, where the baseline cannot perform well even with a larger number of training shapes, while we perform similarly across datasets. 
\begin{table}[]
\parbox{.58\linewidth}{
\centering
\caption{Generalization. Networks trained on the \emph{Cabinet} category, but evaluated on every category of ShapeNet, as well as on Dynamic FAUST. We significantly outperform the baseline (BL) and DeepSDF (DSDF) almost everywhere.}
\setlength\tabcolsep{3pt} %
\resizebox{0.58\columnwidth}{!}{
\begin{tabular}{l|ccc|ccc|ccc} 
\hline
 & \multicolumn{3}{c}{IoU}  & \multicolumn{3}{c}{Chamfer} & \multicolumn{3}{c}{F-score} \\
Category                  &  BL & DSDF & Ours      &  BL & DSDF & Ours  & BL & DSDF & Ours \\
\hline
airplane                  &  33.5 & 56.9 & \textbf{88.2} &         0.668 & 0.583 & \textbf{0.005} &   33.5 & 61.7 & \textbf{96.3}\\
bench                     &  49.1 & 58.8 & \textbf{80.4} &         0.169 & 0.093 & \textbf{0.006} &   63.6 & 76.3 & \textbf{93.3} \\
\emph{cabinet}            &  86.0 & 91.1 & \textbf{91.4} &         0.045 & \textbf{0.025} & 0.121 &   86.4 & \textbf{92.6} & 91.7\\
car                       &  78.4 & 83.7 & \textbf{92.0} &         0.101 & 0.074 & \textbf{0.050} &   62.7 & 73.9 & \textbf{87.2}\\
chair                     &  50.7 & 61.8 & \textbf{86.9} &         0.473 & 0.287 & \textbf{0.012} &   49.1 & 65.2 & \textbf{92.5}\\
display                   &  83.2 & 87.6 & \textbf{94.4} &         0.111 & 0.065 & \textbf{0.052} &   83.9 & 89.6 & \textbf{96.9}\\
lamp                      &  49.7 & 59.3 & \textbf{86.6} &         0.689 & 2.645 & \textbf{0.082} &   50.4 & 64.5 & \textbf{93.4} \\
rifle                     &  56.4 & 56.1 & \textbf{91.8} &         0.114 & 2.669 & \textbf{0.002} &   71.0 & 54.7 & \textbf{99.1}\\
sofa                      &  81.1 & 87.3 & \textbf{94.8} &         0.245 & 0.193 & \textbf{0.010} &   74.2 & 84.6 & \textbf{95.2}\\
speaker                   &  83.2 & 88.3 & \textbf{90.5} &         0.163 & \textbf{0.080} & 0.232 &   71.8 & 80.1 & \textbf{84.9}\\
table                     &  55.0 & 73.6 & \textbf{88.4} &         0.469 & 0.222 & \textbf{0.020} &   61.8 & 82.8 & \textbf{95.0} \\
telephone                 &  90.4 & 94.7 & \textbf{97.3} &         0.051 & 0.015 & \textbf{0.004} &   90.8 & 96.1 & \textbf{99.2}\\
watercraft                &  66.5 & 73.5 & \textbf{91.8} &         0.115 & 0.157 & \textbf{0.006} &   63.0 & 74.2 & \textbf{96.2} \\\hline
mean                      &  66.4 & 74.8 & \textbf{90.3} &         0.263 & 0.547 & \textbf{0.046} &   66.3 & 76.6 & \textbf{93.9}  \\\hline\hline
DFAUST                    &  57.8 & 71.2 & \textbf{94.4} &         0.751 & 0.389 & \textbf{0.012} &   25.0 & 45.4 & \textbf{94.0}\\
\hline
\end{tabular}
\label{tab:generalizationCategory}
} %
}
\hfill
\parbox{.40\linewidth}{
\centering
\caption{Ablative Analysis. We evaluate the performance using different numbers of patches, as well as using variable sizes of the patch latent code/hidden dimensions, and the training data. The training time is measured on an Nvidia V100 GPU.}
\setlength\tabcolsep{3pt} %
\resizebox{0.42\columnwidth}{!}{
\begin{tabular}{l|cccc} 
\cline{2-5}
               &     IoU  & Chamfer  & F-score  &  Time \\\hline
  $N_P=3$     &    73.8  & 0.15    & 72.9  & 1h\\ 
  $N_P=10$     &    85.2  & 0.049    & 88.0  & 1.5h\\ 
  size 32     &    82.8  & 0.066    & 84.7  & 1.5h\\ 
  size 512    &    95.3  & 0.048    & 97.2  & 8h\\ 
  full dataset &    92.2  & 0.050    & 94.8  & 156h\\\hline
  ours         &    91.6  & 0.045    & 94.5  & 2h\\ 
\hline
\end{tabular}
\label{table:ablative}
}
}
\end{table}

\subsubsection{Ablation Experiments}\label{sec:main_ablation}
We perform several ablative analysis experiments to evaluate our approach.
We first evaluate the number of patches required to reconstruct surfaces. 
Table~\ref{table:ablative} reports these numbers on the reduced test set. 
The patch networks here are trained on the reduced training set, consisting of $100$ shapes per ShapeNet category. 
As expected, the performance becomes better with a larger number of patches, since this would lead to smaller patches which can capture more details and generalize better.
We also evaluate the impact of different sizes of the latent codes and hidden dimensions used for the patch network. 
Larger latent codes and hidden dimensions lead to higher quality results. 
Similarly, training on the full training dataset, consisting of $33k$ shapes leads to higher quality. 
However, all design choices with better performance come at the cost of longer training times, see Table~\ref{table:ablative}.

\subsection{Object-Level Priors}\label{sec:objectprior}

We also experiment with \emph{category-specific} object priors.
We add ObjectNet (four FC layers with hidden dimension 1024 and ReLU activations) in front of PatchNet and our baselines.
From object latent codes of size 256, ObjectNet regresses patch latent codes and extrinsics as an intermediate representation usable with PatchNet.
ObjectNet effectively increases the network capacity of our baselines.

\subsubsection{Training}
We initialize all object latents with zeros and the weights of ObjectNet's last layer with very small numbers. 
We initialize the bias of ObjectNet's last layer with zeros for patch latent codes and with the extrinsics of an arbitrary object from the category as computed by our initialization in Sec.~\ref{sec:prelim}.
We pretrain PatchNet on ShapeNet. 
For our method, the PatchNet is kept fixed from this point on.
As training set, we use the full training split of the ShapeNet category for which we train.
We remove $\mathcal{L}_\text{rot}$ completely as it significantly lowers quality. 
The $L2$ regularization is only applied to the object latent codes. 
We set $\omega_\text{var}=5$.
ObjectNet is trained  in three phases, each lasting 1000 epochs. 
We use the same initial learning rates as when training PatchNet, except in the last phase, where we reduce them by a factor of 5.
The batch size is 128.\\
\emph{Phase I}: We pretrain ObjectNet to ensure good patch extrinsics.
For this, we use the extrinsic loss, $\mathcal{L}_\text{ext}$ in Eq.~\ref{eq:loss}, and the regularizer. We set $\omega_\text{scl}=2$.\\
\emph{Phase II}: Next, we learn to regress patch latent codes. First, we add a layer that multiplies the regressed scales by 1.3. We then store these extrinsics. Afterwards, we train using $\mathcal{L}_\text{recon}$ and two $L2$ losses that keep the regressed position and scale close to the stored extrinsics, with respective weights $1,3,$ and $30$.\\
\emph{Phase III}: The complete loss $\mathcal{L}$ in Eq.~\ref{eq:loss_main}, with $\omega_\text{scl}=0.02$, yields final refinements.

\subsubsection{Coarse Correspondences} 
\begin{figure}
\centering

\includegraphics[trim={240 70 130 150},clip,height=1.2cm]
{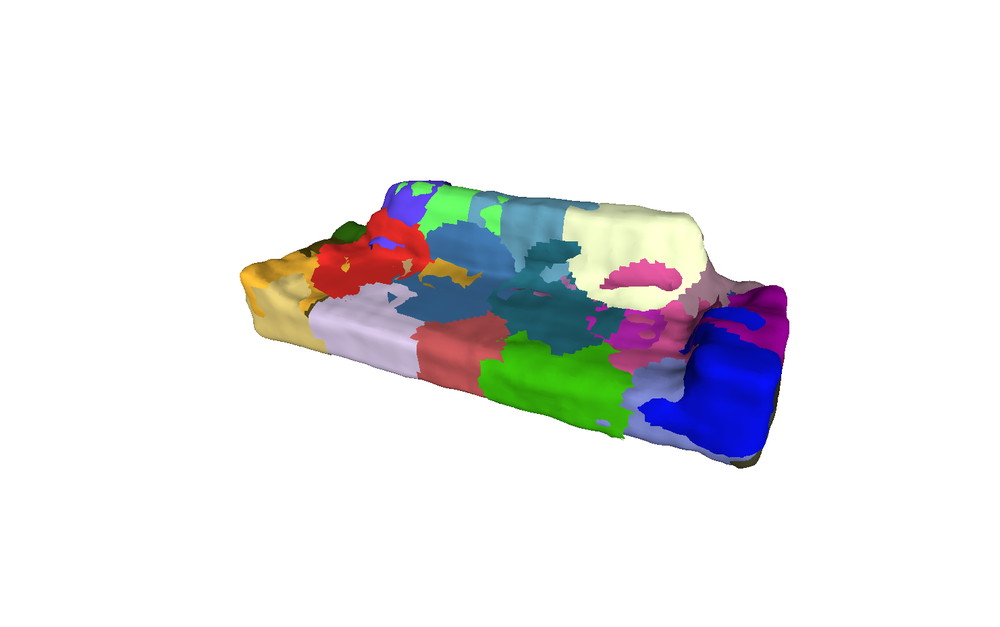}
\includegraphics[trim={220 70 170 130},clip,height=1.2cm]
{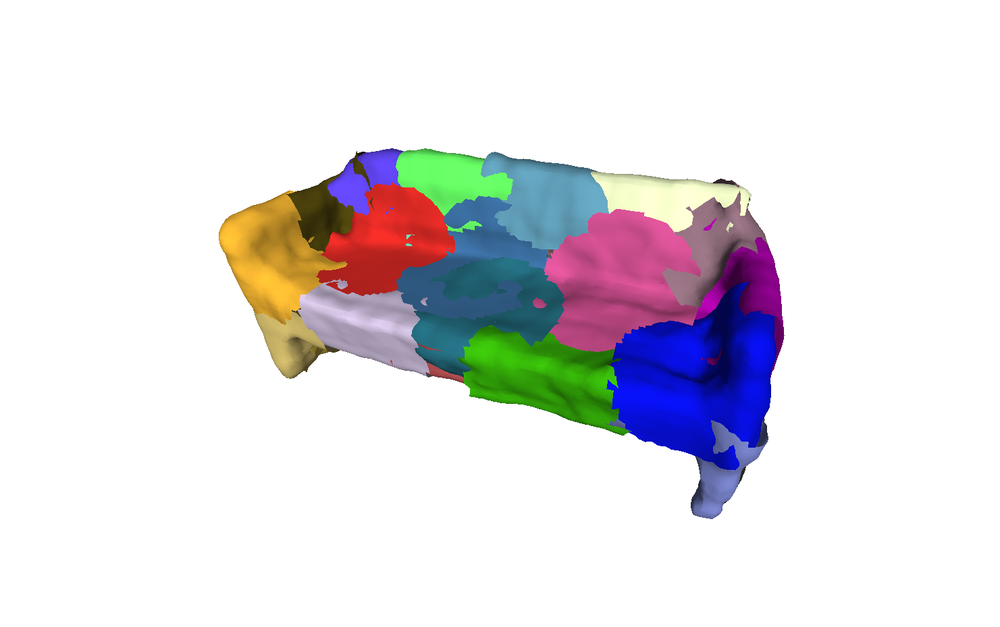}
\includegraphics[trim={255 70 220 110},clip,height=1.2cm]
{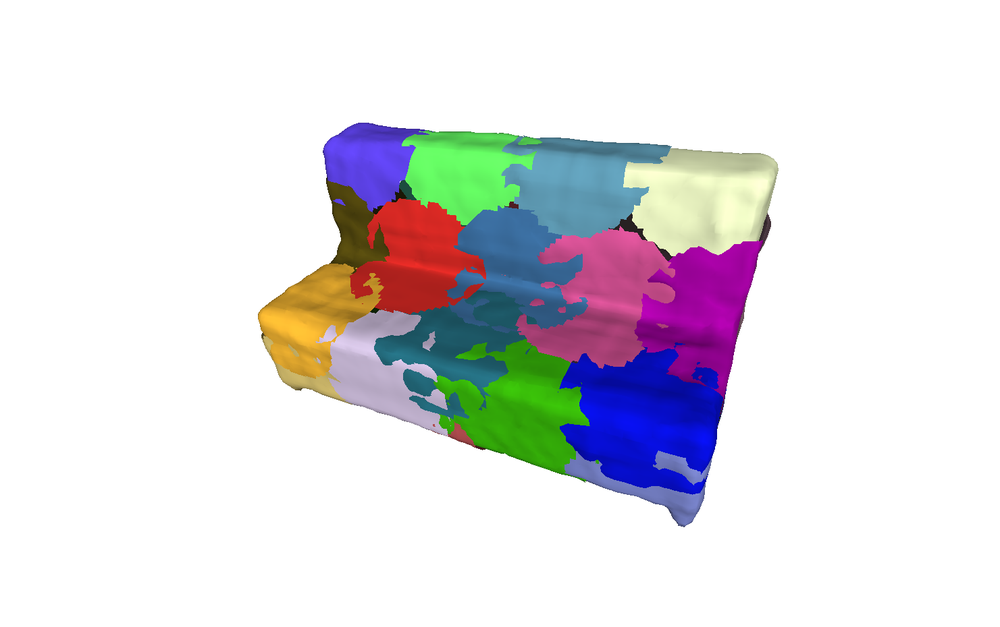}
\includegraphics[trim={220 70 170 150},clip,height=1.2cm]
{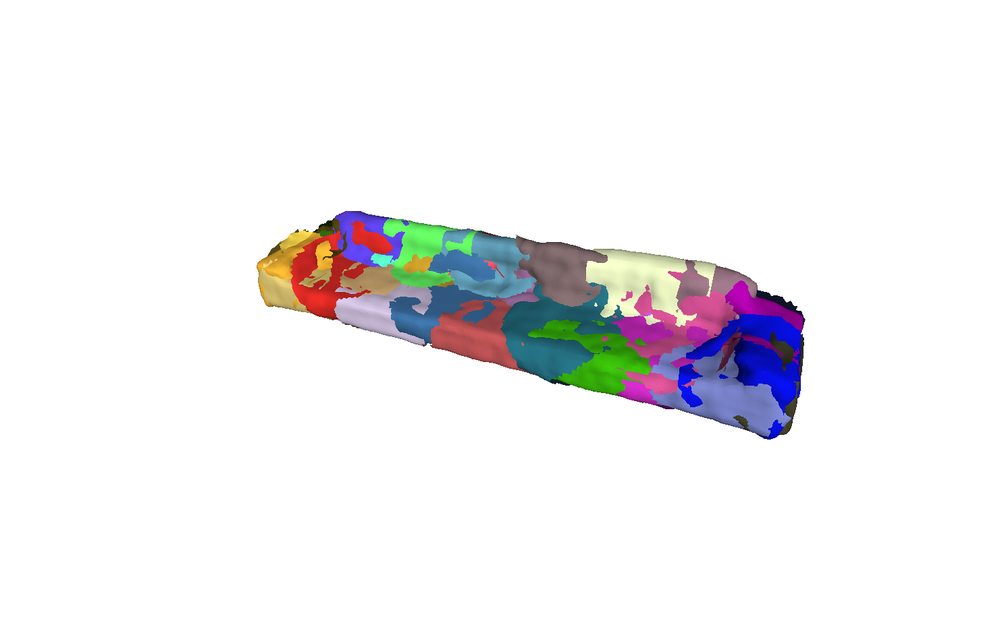}
\includegraphics[trim={220 70 210 160},clip,height=1.2cm]
{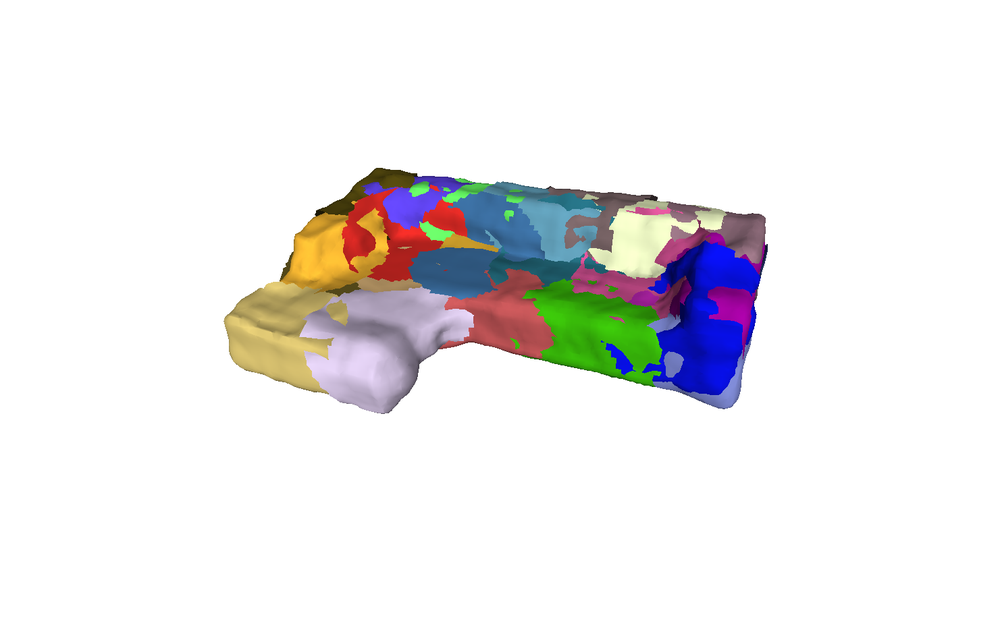}
\includegraphics[trim={240 70 240 120},clip,height=1.2cm]
{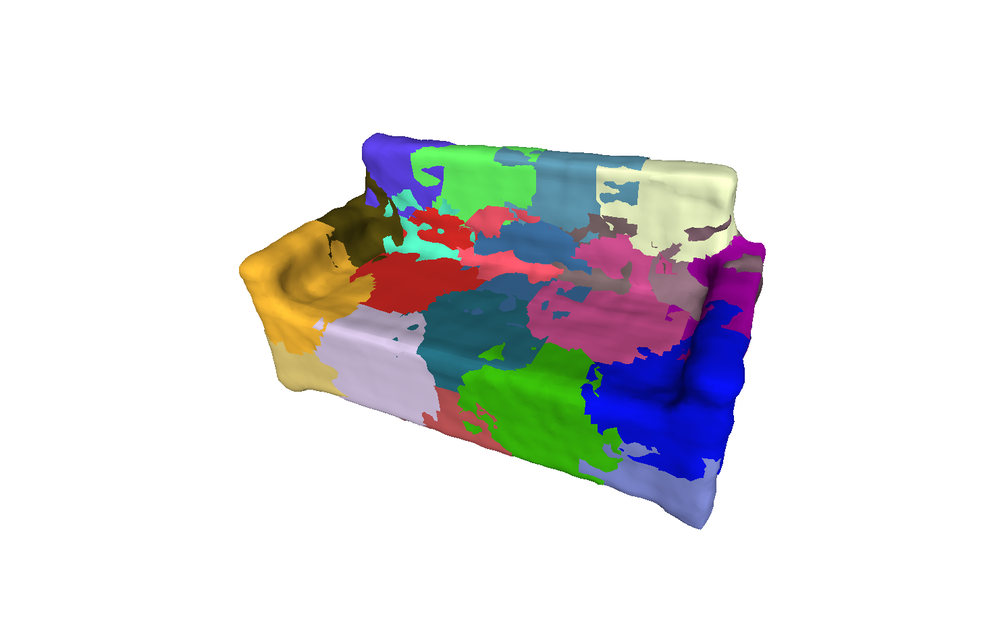}
\includegraphics[trim={235 70 200 150},clip,height=1.2cm]
{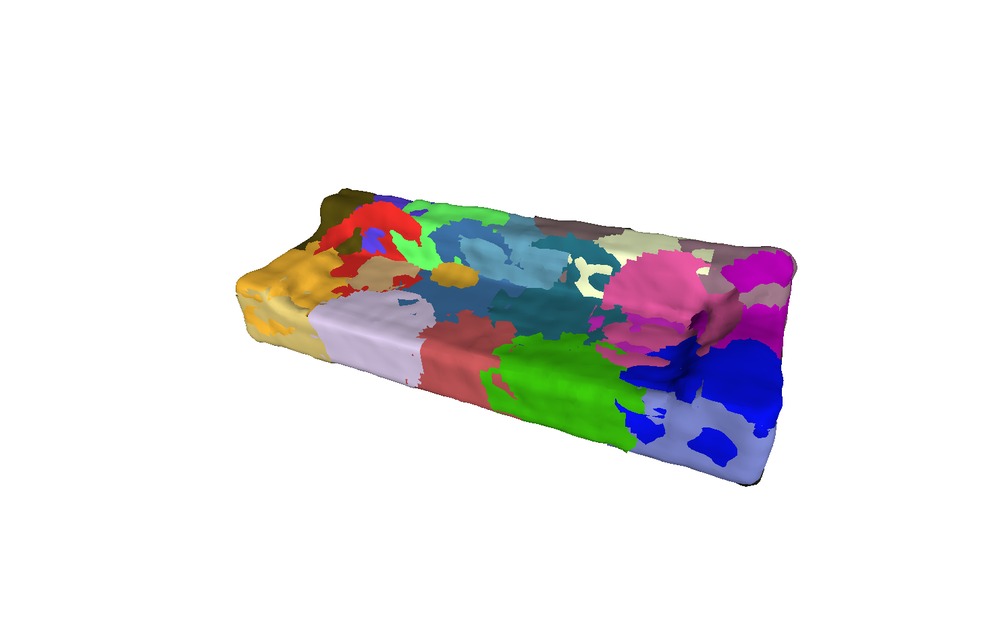}
\caption{Coarse Correspondences. Note the consistent coloring of the patches.}
\label{fig:corresondences}
\end{figure}

Fig.~\ref{fig:corresondences} shows that the learned patch distribution is consistent across objects, establishing coarse correspondences between objects.

\subsubsection{Interpolation}

Due to the implicitly learned coarse correspondences, we can encode test objects into object latent codes and then linearly interpolate between them.
Fig.~\ref{fig:generative} shows that interpolation of the latent codes leads to a smooth morph between the decoded shapes in 3D space.

\subsubsection{Generative Model}

\begin{figure}
\centering

\includegraphics[trim={180 70 160 100},clip,height=1.5cm]
{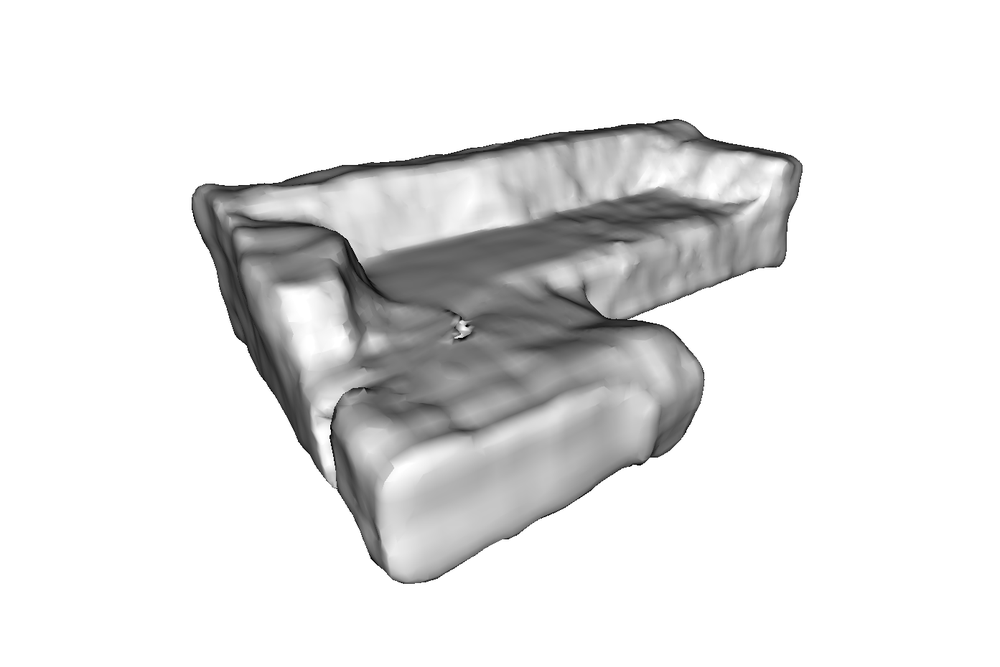}
\includegraphics[trim={180 70 160 100},clip,height=1.5cm]
{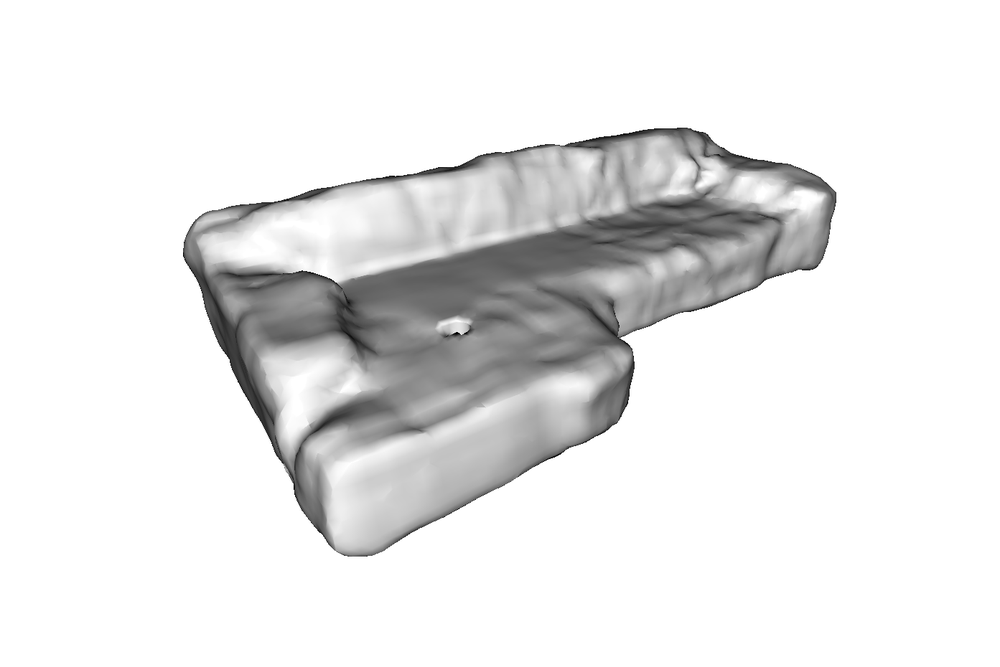}
\includegraphics[trim={180 70 130 100},clip,height=1.5cm]
{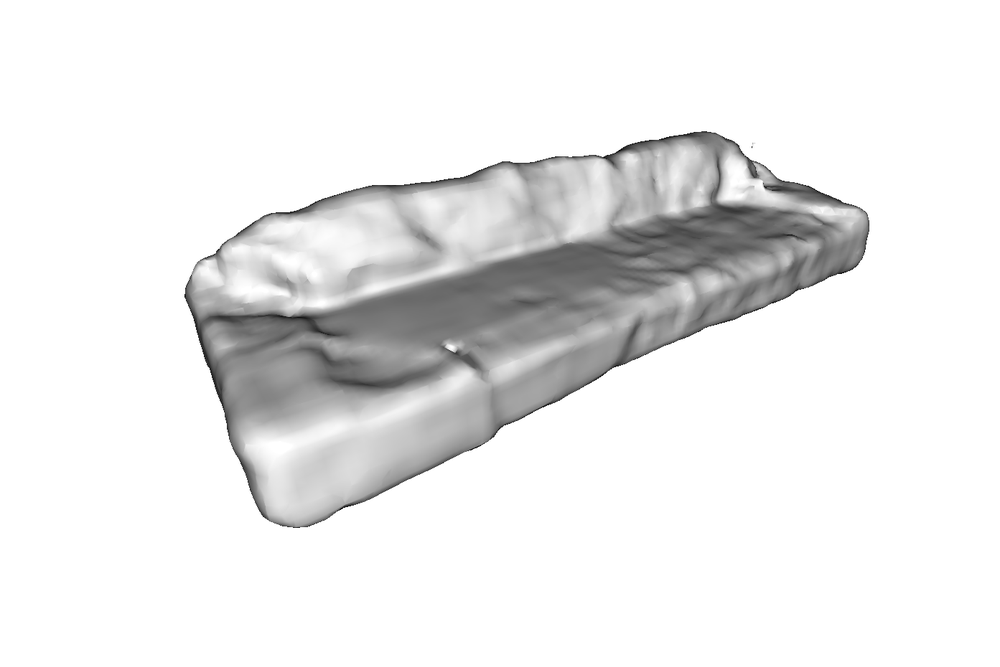}
\includegraphics[trim={180 70 130 100},clip,height=1.5cm]
{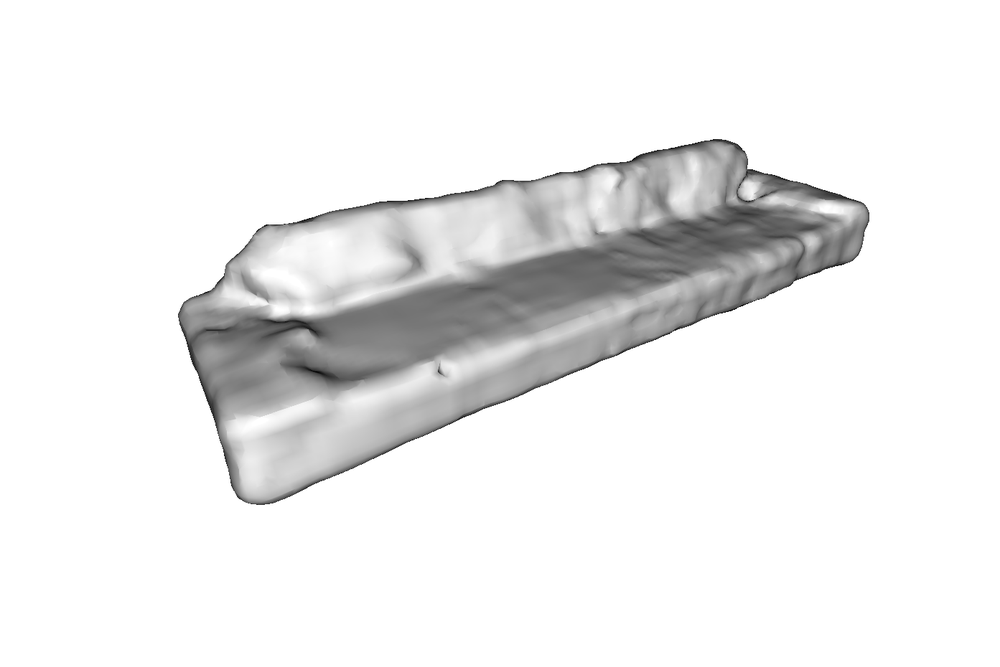}
\includegraphics[trim={180 70 130 100},clip,height=1.5cm]
{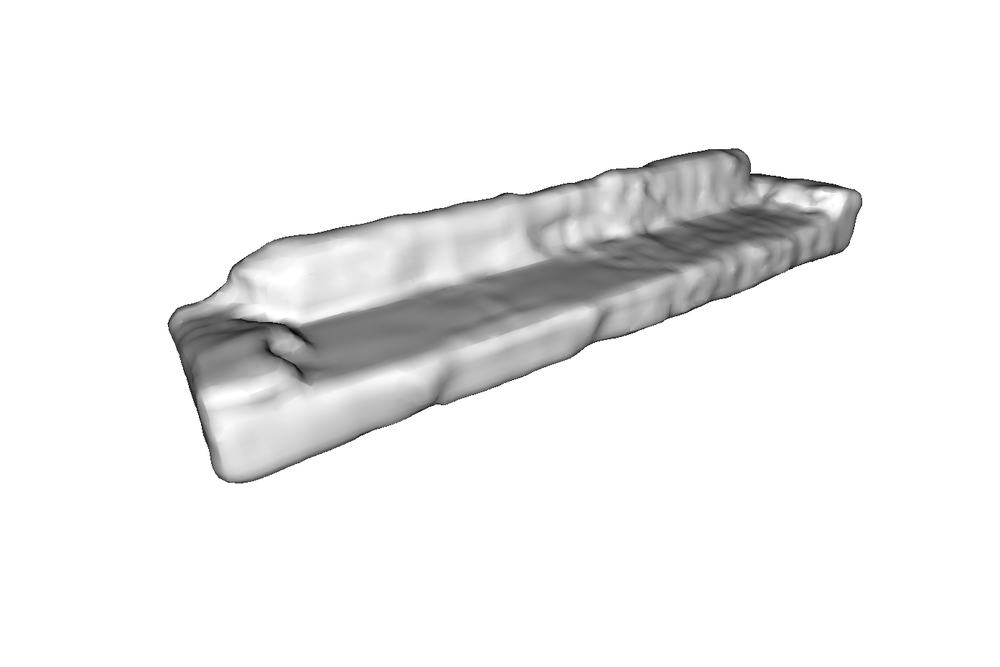}

\includegraphics[trim={270 100 240 200},clip,height=1.3cm]
{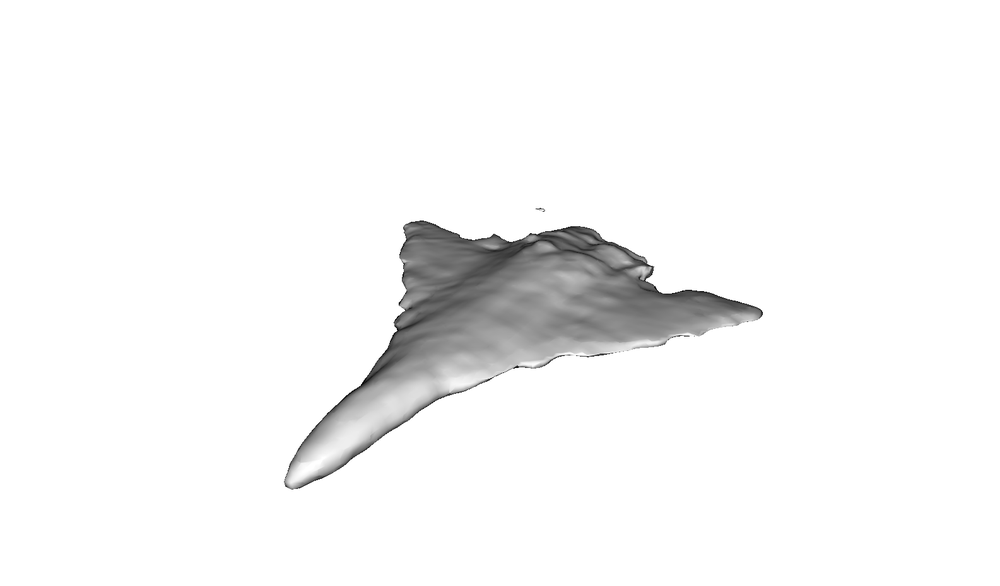}
\includegraphics[trim={260 0 170 150},clip,height=1.3cm]
{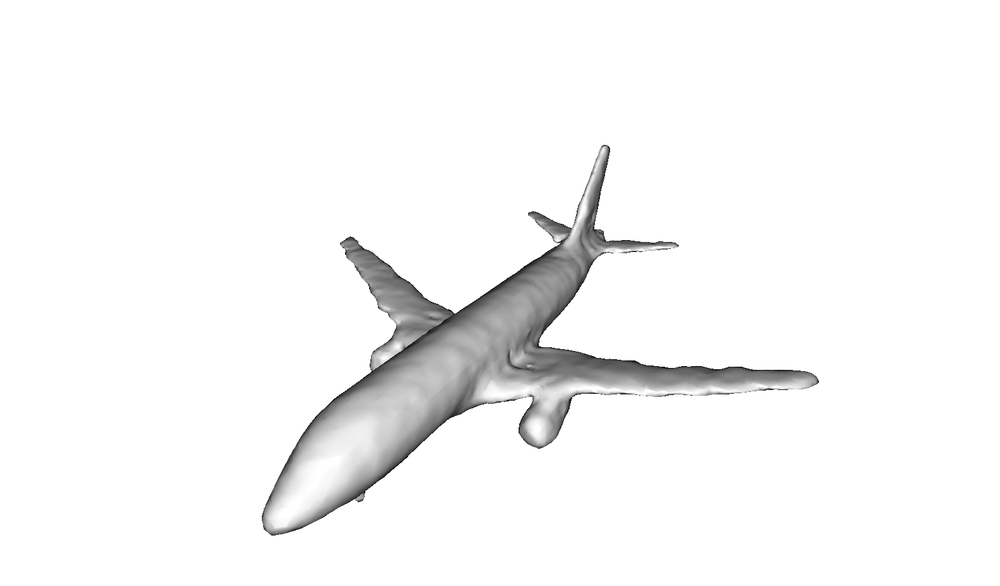}
\includegraphics[trim={270 60 220 200},clip,height=1.3cm]
{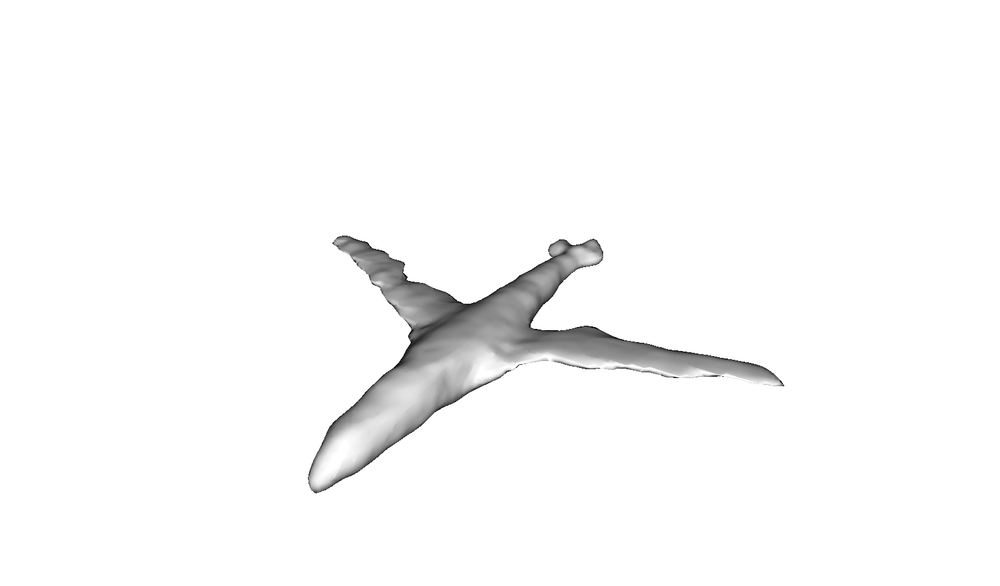}
\includegraphics[trim={240 120 200 120},clip,height=1.3cm]
{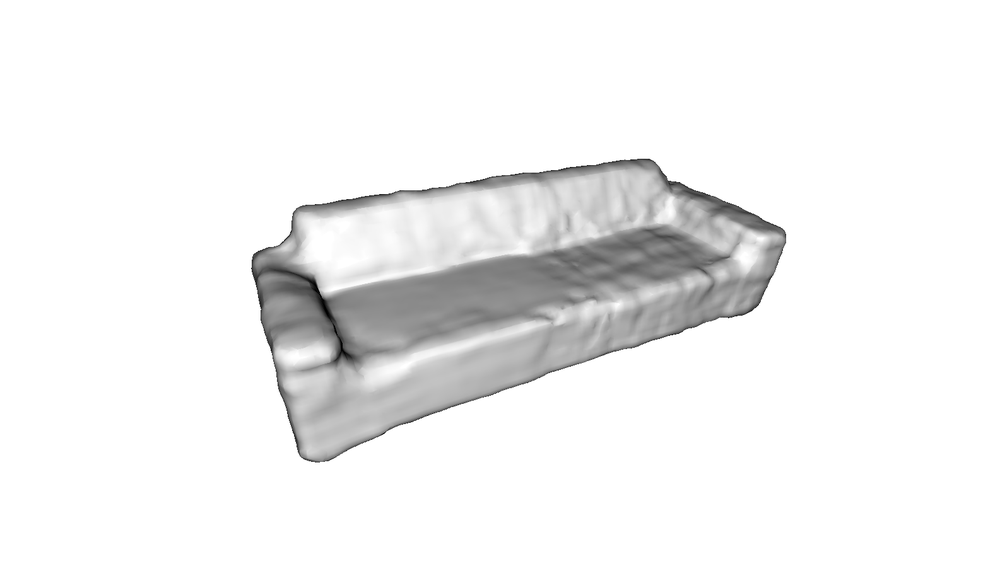}
\includegraphics[trim={270 120 220 170},clip,height=1.3cm]
{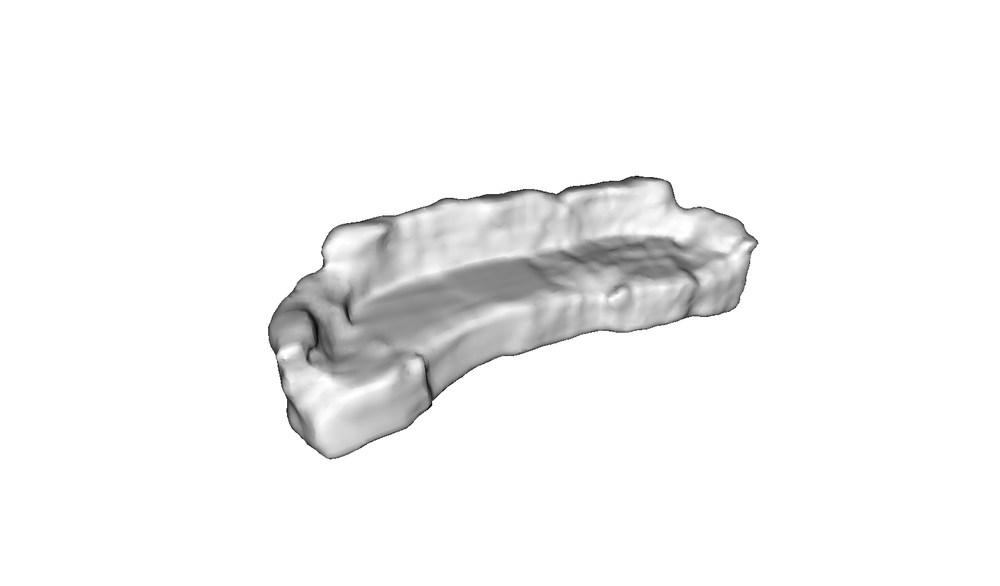}

\caption{Interpolation (top). The left and right end points are encoded test objects. Generative Models (bottom). We sample object latents from ObjectNet's fitted prior.}
\label{fig:generative}
\end{figure}

We can explore the learned object latent space further by turning ObjectNet into a generative model.
Since auto-decoding does not yield an encoder that inputs a known distribution, we have to estimate the unknown input distribution.
Therefore, we fit a multivariate Gaussian to the object latent codes obtained at training time.
We can then sample new object latent codes from the fitted Gaussian and use them to generate new objects, see Fig.~\ref{fig:generative}.

\subsubsection{Partial Point Cloud Completion}
Given a partial point cloud, we can optimize for the object latent code which best explains the visible region. 
ObjectNet acts as a prior which completes the missing parts of the shape. 
For our method, we pretrained our PatchNet on a different object category and keep it fixed, and then train ObjectNet on the target category, which makes this task more challenging for us.
We choose the versions of our baselines where the eight final layers are pretrained on all categories and finetuned on the target shape category.
We evaluated several other settings, with this one being the most competitive. 
See the supplemental for more on surface reconstruction with object-level priors.

\begin{figure}
\centering

\includegraphics[trim={180 0 100 0},clip,height=1.7cm]
{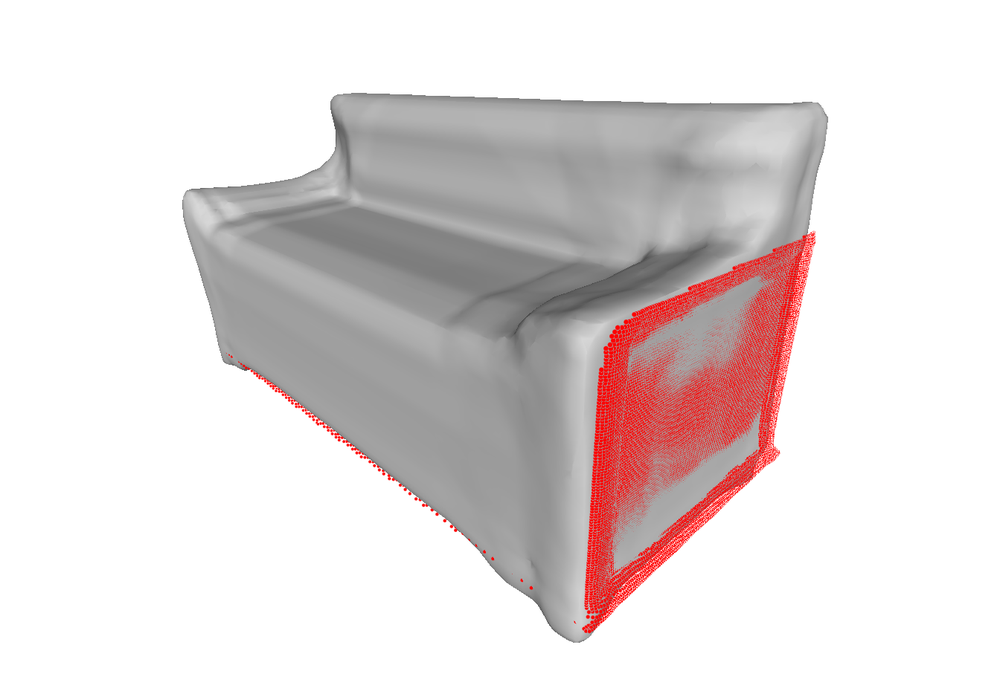}
\includegraphics[trim={140 0 100 0},clip,height=1.7cm]
{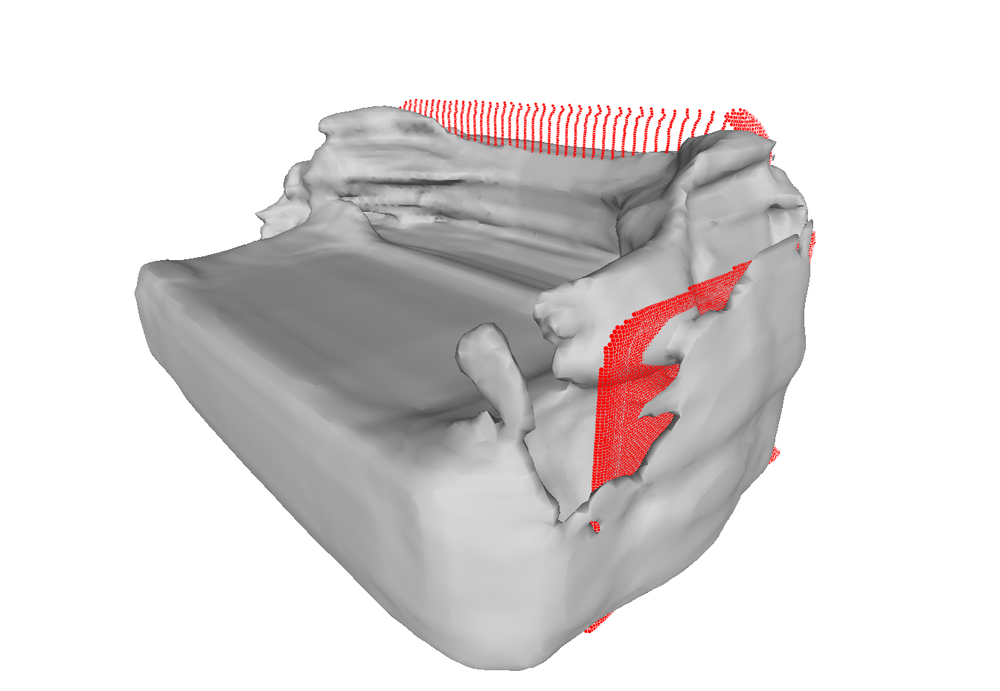}
\includegraphics[trim={180 0 100 0},clip,height=1.7cm]
{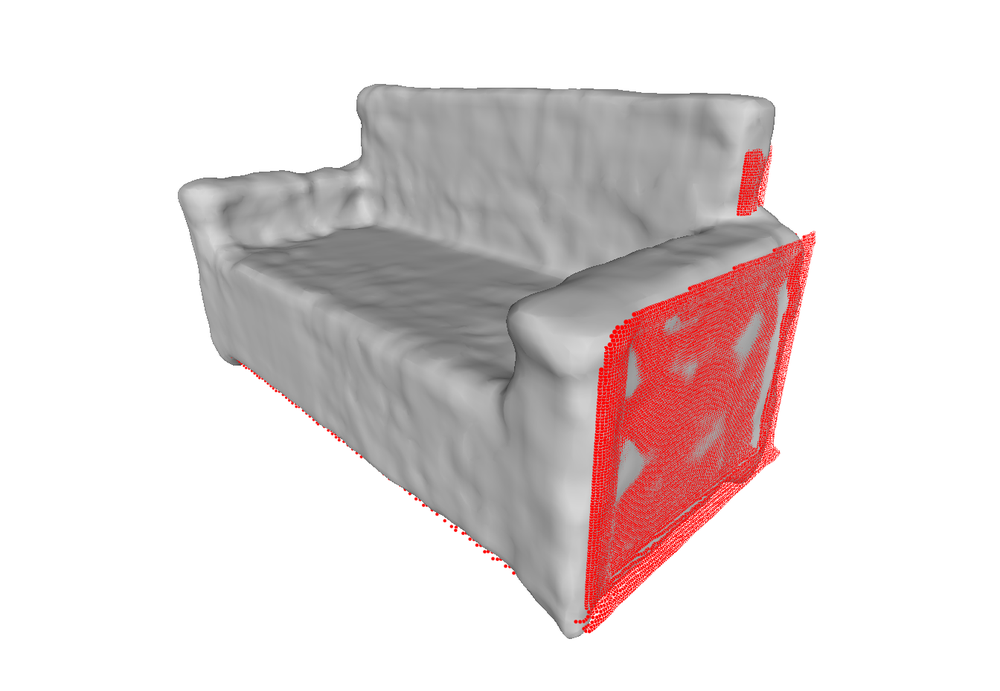}
\includegraphics[trim={180 0 100 0},clip,height=1.7cm]
{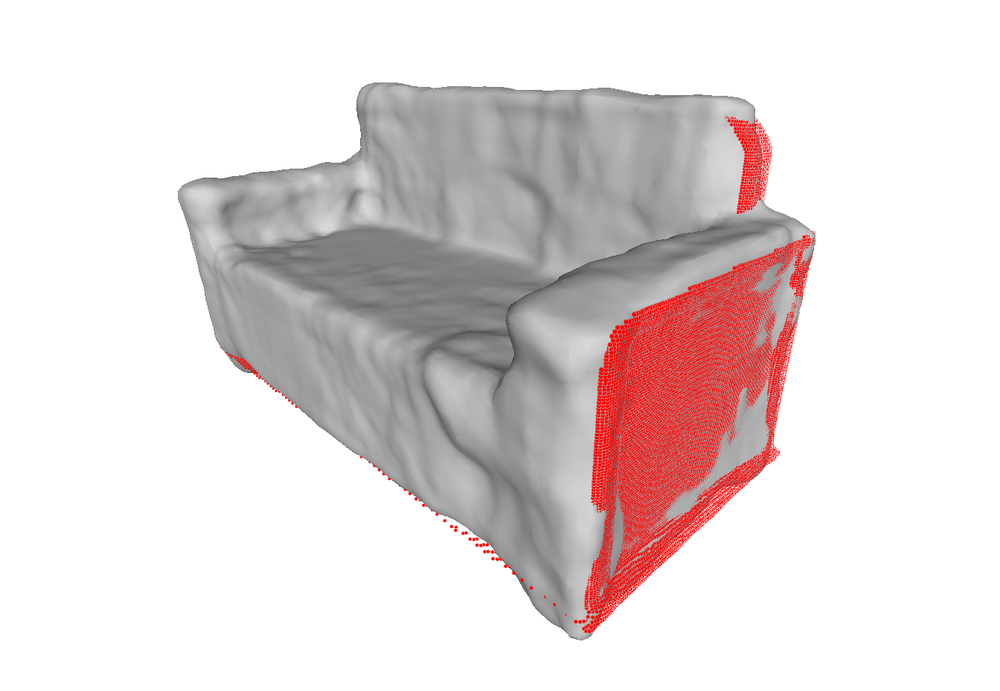}
\hfill
\includegraphics[trim={220 100 70 0},clip,height=1.7cm]
{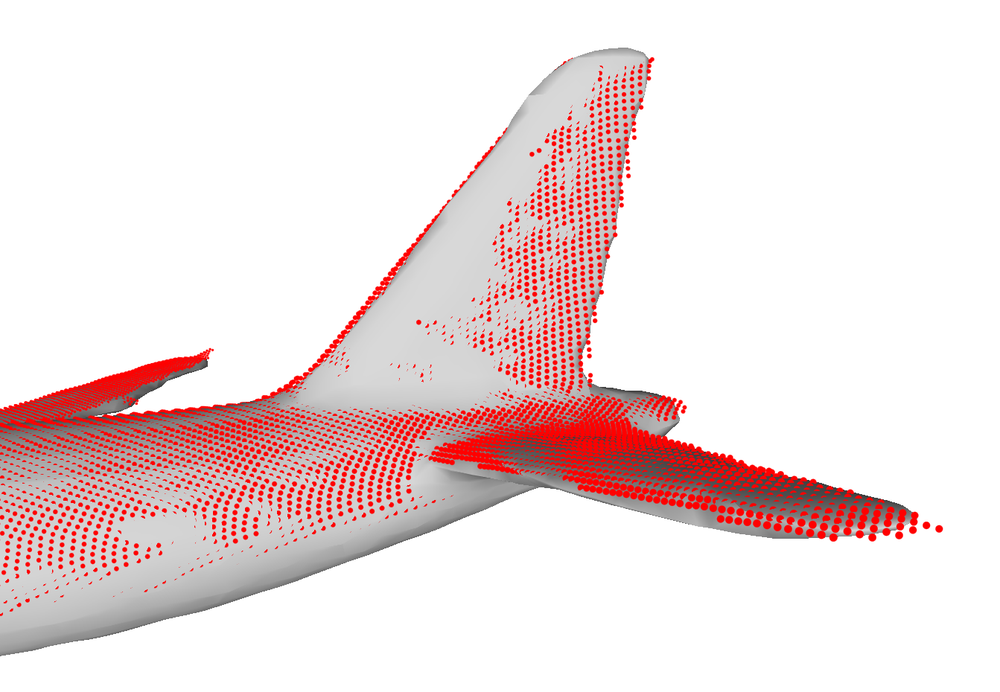}
\includegraphics[trim={220 100 70 0},clip,height=1.7cm]
{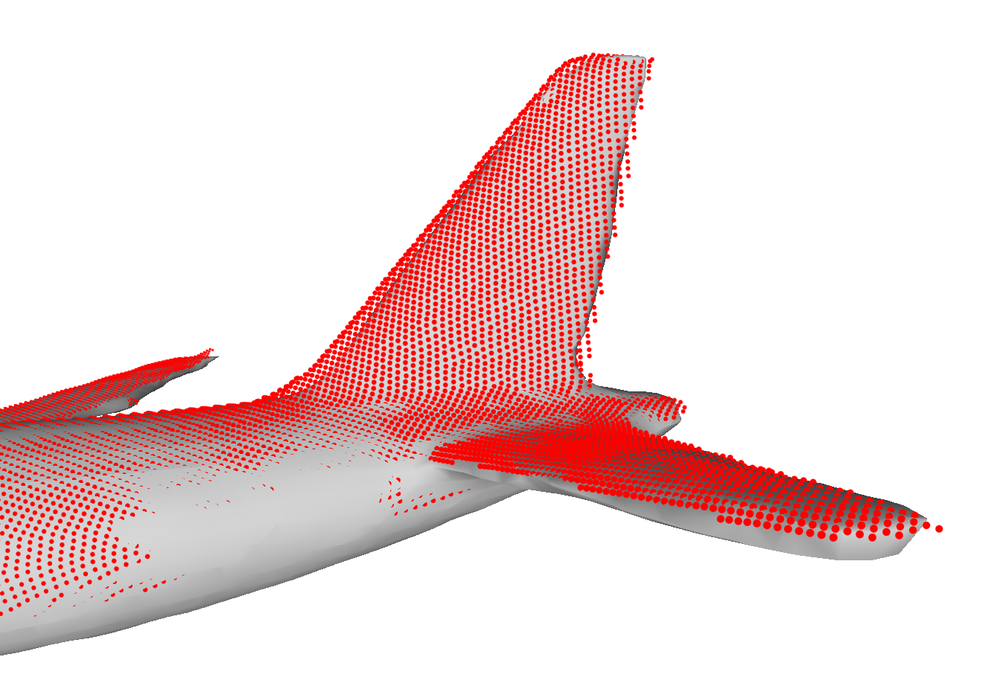}

\caption{Shape Completion. (Sofa) from left to right: Baseline, DeepSDF, ours unrefined, ours refined. (Airplane) from left to right: Ours unrefined, ours refined.}
\label{fig:completion}
\end{figure}

\emph{Optimization}: We initialize with the average of the object latent codes obtained at training time. %
We optimize for 600 iterations, starting with a learning rate of $0.01$ and halving it every 200 iterations.
Since our method regresses the patch latent codes and extrinsics as an intermediate step, we can further refine the result by treating this intermediate patch-level representation as free variables. 
Specifically, we refine the patch latent code for the last 100 iterations with a learning rate of $0.001$, while keeping the extrinsics fixed. This allows to integrate details not captured by the object-level prior.
Fig.~\ref{fig:completion} demonstrates this effect.
During optimization, we use the reconstruction loss, the $L2$ regularizer and the coverage loss. 
The other extrinsics losses have a detrimental effect on patches that are outside the partial point cloud.
We use 8k samples per iteration.

We obtain the partial point clouds from depth maps similar to Park~\etal~\cite{park2019deepsdf}. 
We also employ their free-space loss, which encourages the network to regress positive values for samples between the surface and the camera. We use $30\%$ free-space samples.
We consider depth maps from a fixed and from a per-scene random viewpoint. 
For shape completion, we report the F-score between the full groundtruth mesh and the reconstructed mesh.
Similar to Park~\etal~\cite{park2019deepsdf}, we also compute the mesh accuracy for shape completion. It is the 90th percentile of shortest distances from the surface samples of the reconstructed shape to surface samples of the full groundtruth.
Table~\ref{tab:shape_completion_main} shows how, due to local refinement on the patch level, we outperform the baselines everywhere.

\begin{table}
\centering
\caption{Partial Point Cloud Completion from Depth Maps. We complete depth maps from a fixed camera viewpoint and from per-scene random viewpoints.}
\setlength\tabcolsep{3pt} %
\resizebox{0.8\columnwidth}{!}{
\begin{tabular}{l|c|c|c|c|c|c|c|c}
\cline{2-9}
             & \multicolumn{2}{c|}{sofas fixed}      & \multicolumn{2}{c|}{sofas random}     & \multicolumn{2}{c|}{airplanes fixed}      & \multicolumn{2}{c}{airplanes random}     \\
             & acc.       & F-score       & acc.       & F-score       & acc.       & F-score       & acc.       & F-score       \\\hline
baseline                        & 0.094 & 43.0 & 0.092 & 42.7 & 0.069 & 58.1 & 0.066 & 58.7 \\
DeepSDF-based baseline         & 0.106 & 33.6 & 0.101 & 39.5 & 0.066 & 56.9 & 0.065 & 55.5 \\
ours               & 0.091 & 48.1 & 0.077 & 49.2 & 0.058 & 60.5 & 0.056 & 59.4 \\
ours+refined       & \textbf{0.052} &  \textbf{53.6} & \textbf{0.053} &  \textbf{52.4} & \textbf{0.041} &  \textbf{67.7} & \textbf{0.043} &  \textbf{65.8} \\       \hline 
\end{tabular}}
\label{tab:shape_completion_main}
\end{table}

\subsection{Articulated Deformation}

\begin{figure}
\centering

\includegraphics[trim={340 60 370 10},clip,height=2.7cm]
{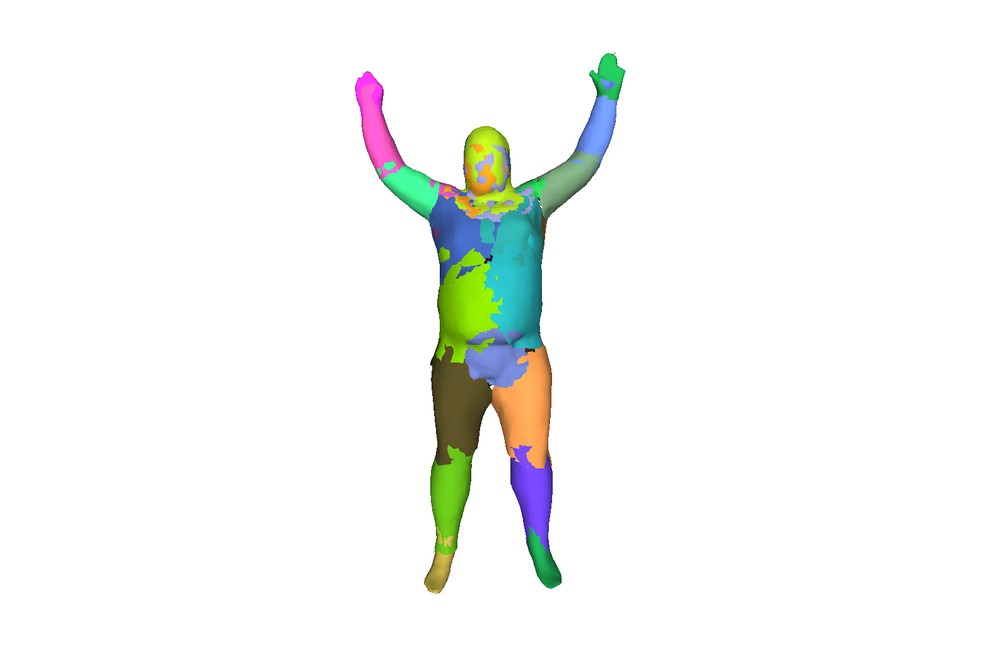}
\includegraphics[trim={320 50 270 0},clip,height=2.7cm]
{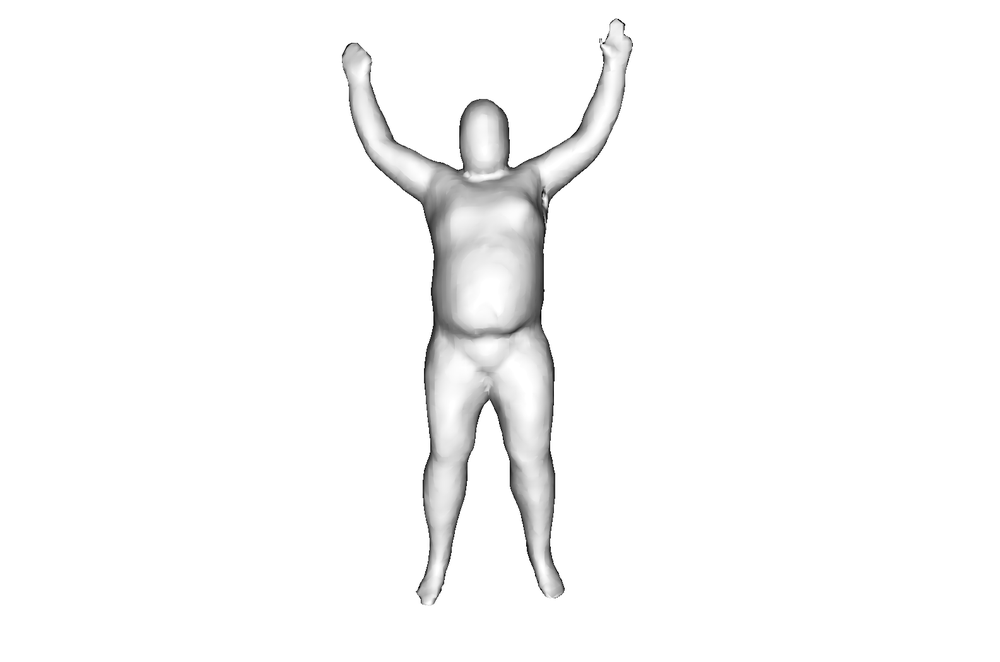}
\includegraphics[trim={380 60 420 20},clip,height=2.7cm]
{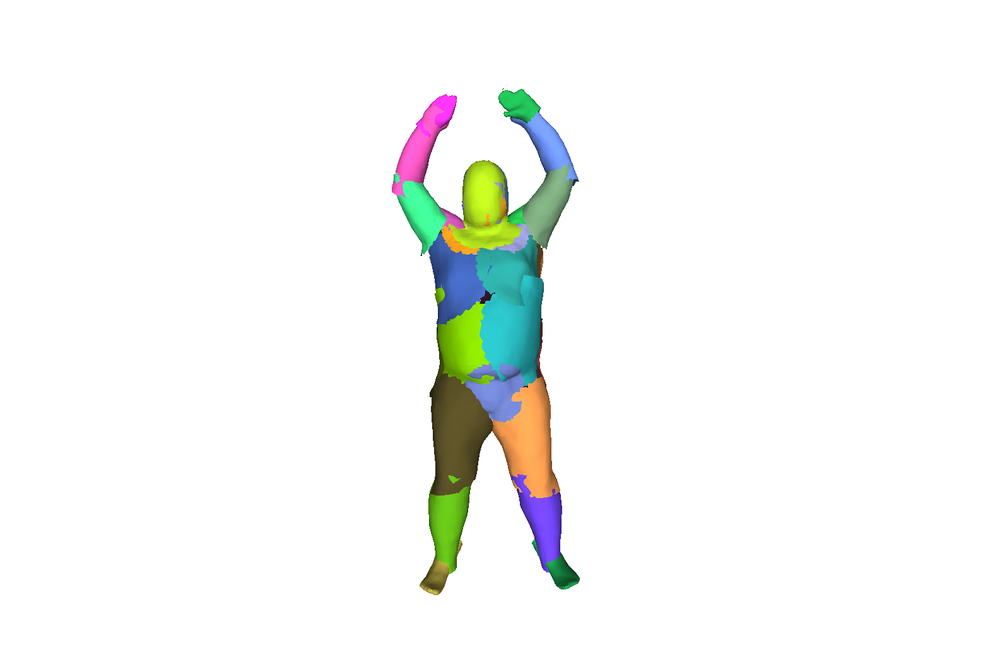}
\includegraphics[trim={380 50 350 0},clip,height=2.7cm]
{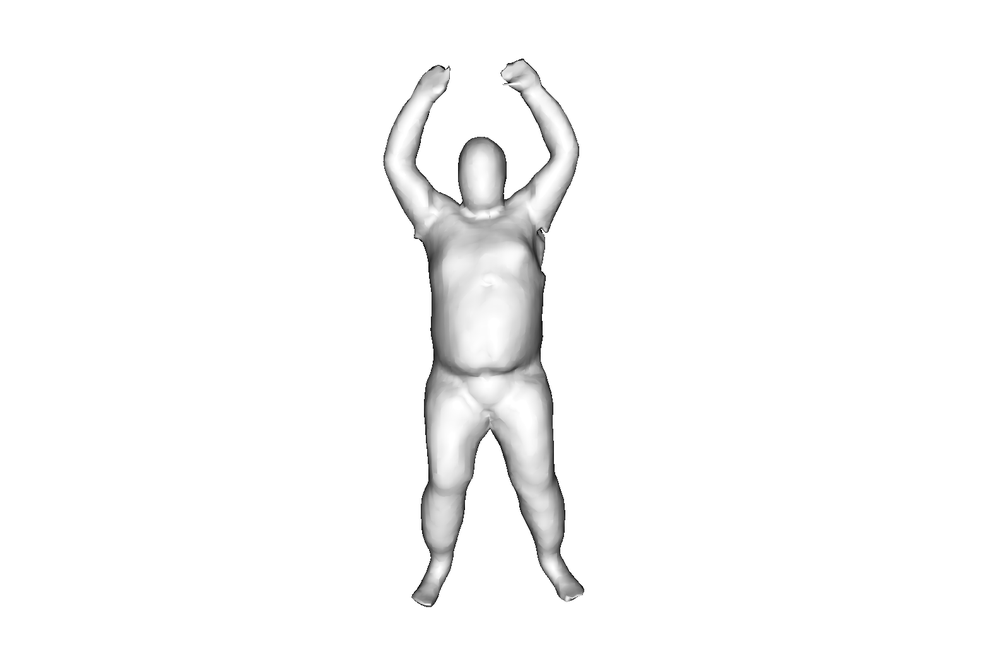}
\includegraphics[trim={290 50 300 30},clip,height=2.7cm]
{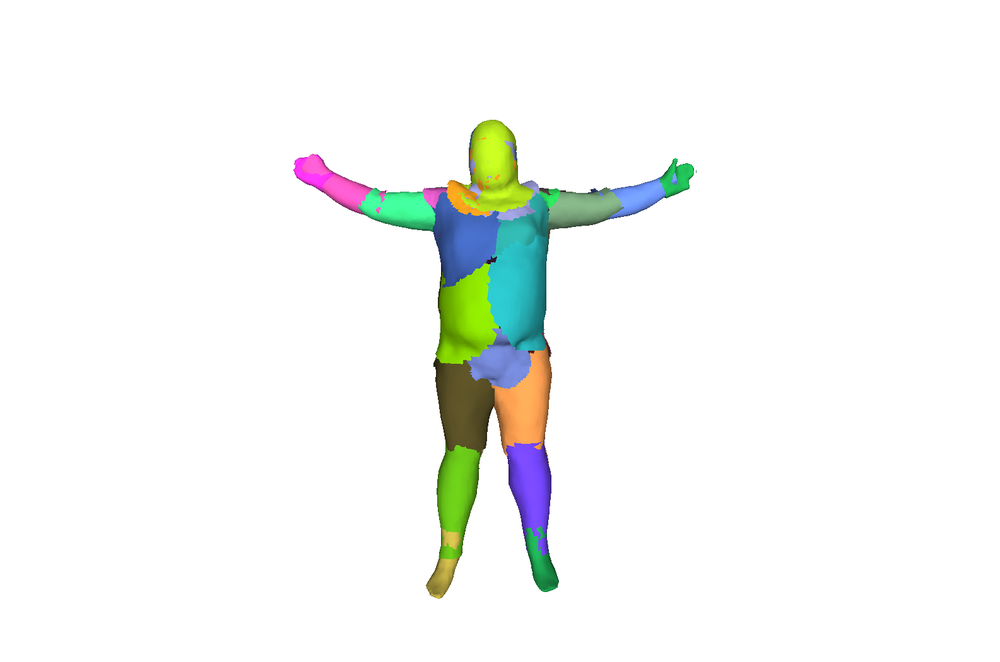}
\includegraphics[trim={270 30 270 0},clip,height=2.7cm]
{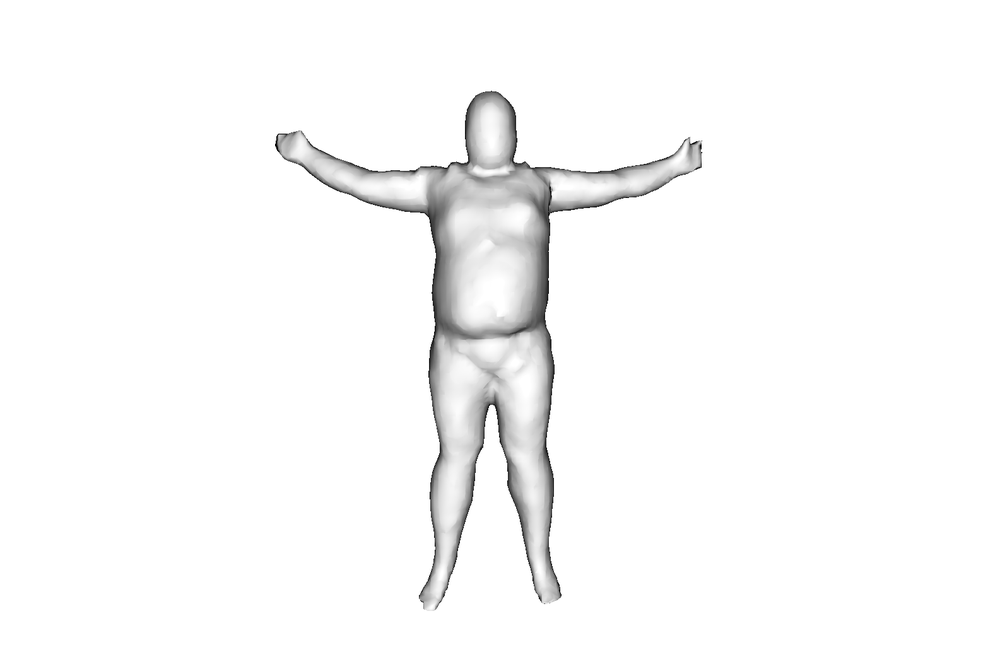}
\includegraphics[trim={400 60 350 30},clip,height=2.7cm]
{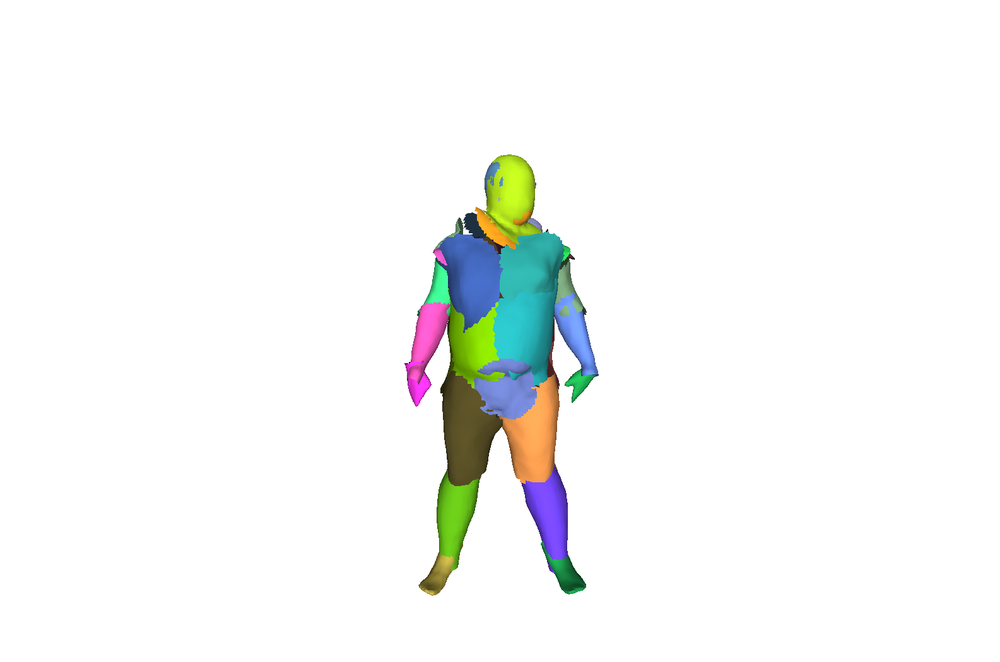}
\includegraphics[trim={400 50 390 0},clip,height=2.7cm]
{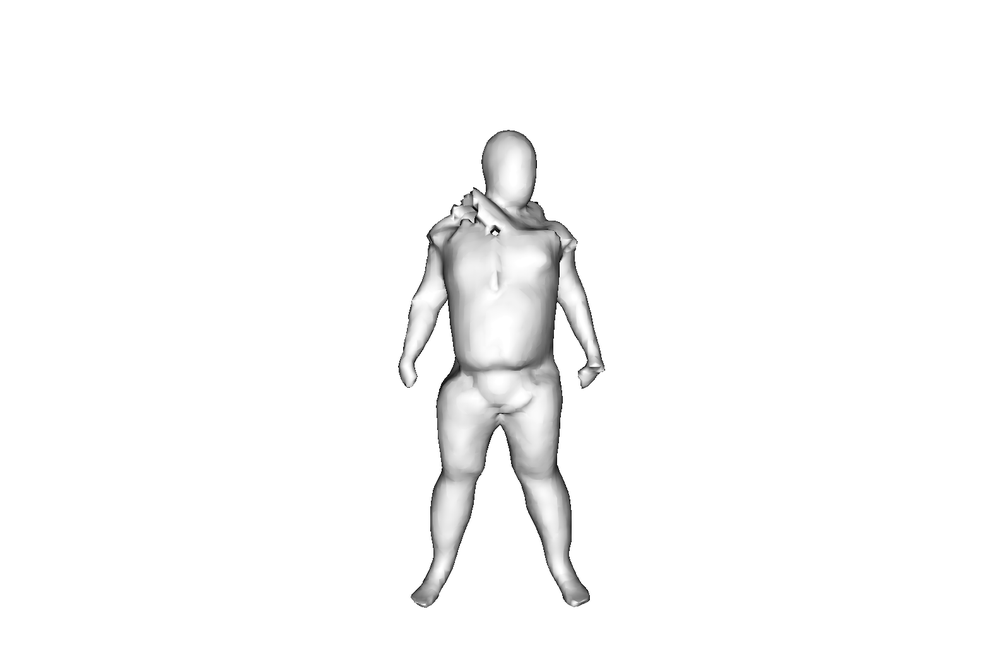}

\caption{Articulated Motion. We encode a template shape into patch latent codes (first pair). We then modify the patch extrinsics, while keeping the patch latent codes fixed, leading to non-rigid deformations (middle two pairs). The last pair shows a failure case due to large non-rigid deformations away from the template. Note that the colored patches move rigidly across poses while the mixture deforms non-rigidly.}
\label{fig:articulated_motion}
\end{figure}

Our patch-level representation can model some articulated deformations by \emph{only} modifying the patch extrinsics, without needing to adapt the patch latent codes. %
Given a template surface and patch extrinsics for this template, we first encode it into patch latent codes.
After manipulating the patch extrinsics, we can obtain an articulated surface with our smooth blending from Eq.~\ref{eq:mixture}, as Fig.~\ref{fig:articulated_motion} demonstrates.

\section{Concluding Remarks}
\noindent\textbf{Limitations.} We sample the SDF using DeepSDF's sampling strategy, 
which might limit the level of detail.
Generalizability at test time requires optimizing patch latent codes and extrinsics, a problem shared with other auto-decoders. %
We fit the reduced test set in 71 min due to batching, one object in 10 min.

\noindent\textbf{Conclusion.} We have presented a mid-level geometry representation based on patches.
This representation leverages the similarities of objects at patch level leading to a highly generalizable neural shape representation.
For example, we show that our representation, trained on one object category can also represent other categories.
We hope that our representation will enable a large variety of applications that go far beyond shape interpolation and point cloud completion.

\noindent\textbf{Acknowledgements.} 
This work was supported by  
the ERC Consolidator Grant 4DReply (770784), 
and an Oculus research grant.

\clearpage
\bibliographystyle{splncs04}
\bibliography{eccv2020kit/sections/egbib}

\clearpage
\title{Supplementary Material}
\titlerunning{PatchNets: Supplementary Material}
\author{}
\institute{}
\maketitle

\setcounter{section}{0}
\renewcommand\thesection{S.\arabic{section}}

\setcounter{figure}{0}
\renewcommand{\thefigure}{S.\arabic{figure}}

\setcounter{table}{0}
\renewcommand{\thetable}{S.\arabic{table}}

In this supplemental material, we expand on some points from the main paper.
We first perform an ablation study on the extrinsics losses in Sec.~\ref{sec:ablation}.
In Sec.~\ref{sec:metrics}, we describe the error measures we employ. 
In Sec.~\ref{sec:reduced}, we compare error measures on the reduced and full test sets.
Sec.~\ref{sec:single} shows the randomly picked single shape that we use in one of the generalization experiments.
Sec.~\ref{sec:object} contains more experiments using object-level priors.
Sec.~\ref{sec:patches} shows different number of patches and network/latent code sizes.
Next, we measure the performance under synthetic noise in Sec.~\ref{sec:noise}.
We show preliminary results on a large scene in Sec.~\ref{sec:large}.
Finally, in Sec.~\ref{sec:dsif}, we provide some remarks on the concurrent work DSIF~\cite{genova2019deep}.

\section{Loss Ablation Study}\label{sec:ablation}
\begin{figure}
\centering

\begin{tabular}{l|c|c}
& mixture & patches\\\hline
$\mathcal{L}_\text{recon}$ on mixture
&
\includegraphics[trim={350 200 200 200},clip,height=2cm]
{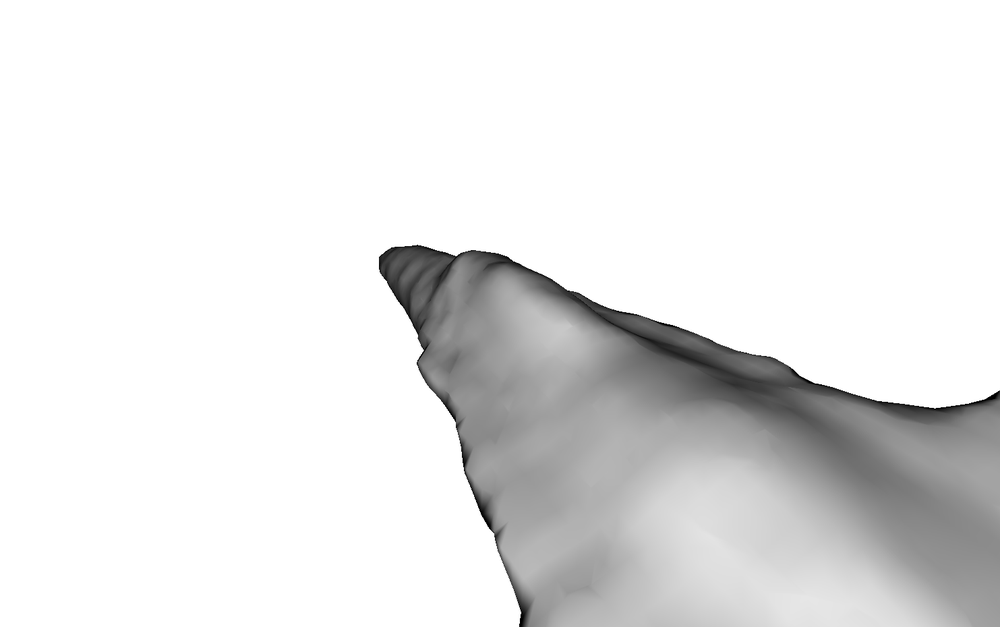}&
\includegraphics[trim={350 200 200 200},clip,height=2cm]
{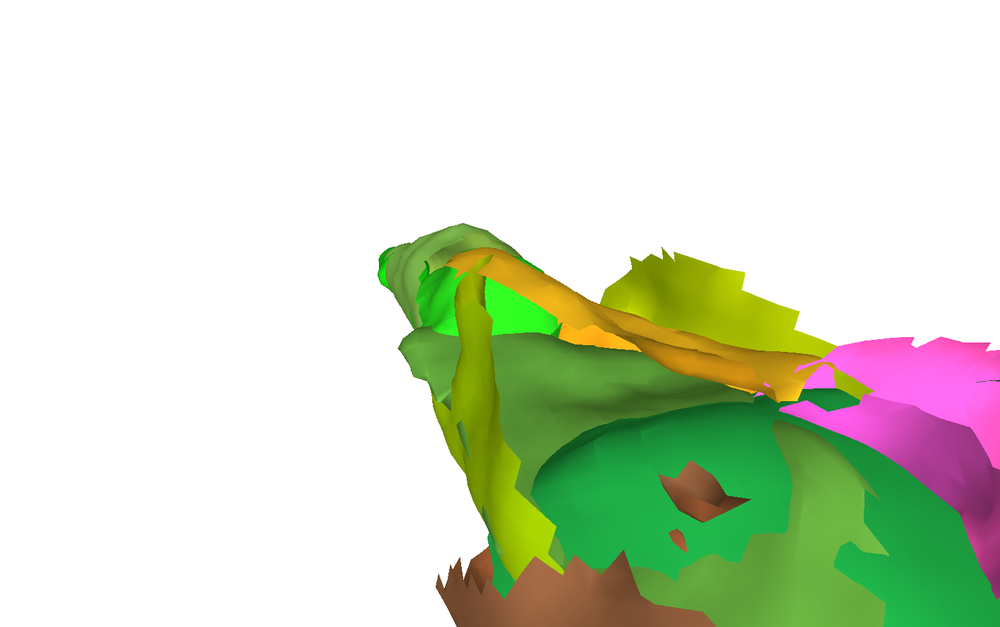}\\\hline
ours
&
\includegraphics[trim={350 200 200 200},clip,height=2cm]
{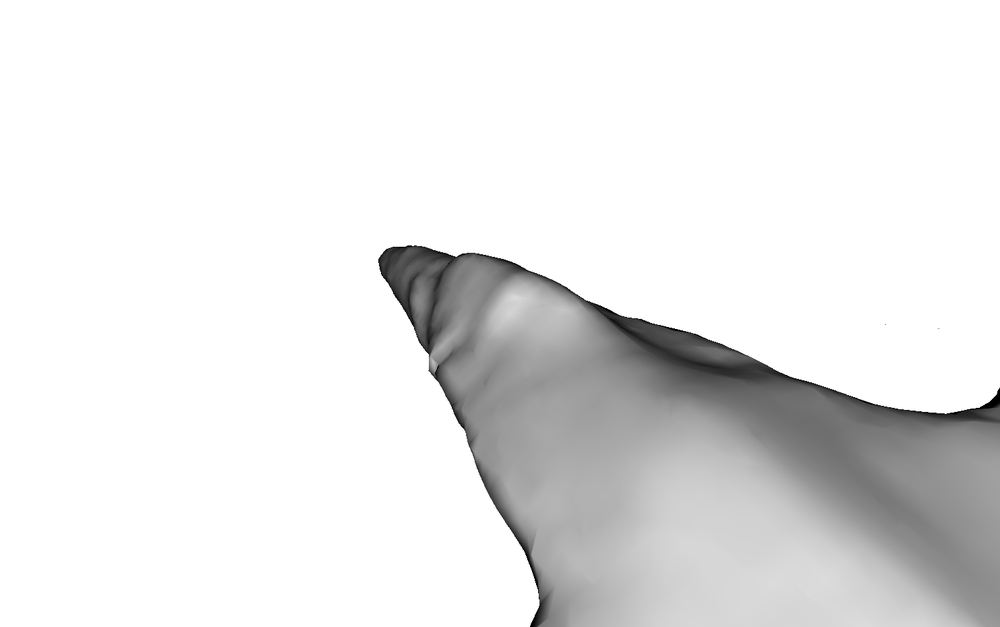}&
\includegraphics[trim={350 200 200 200},clip,height=2cm]
{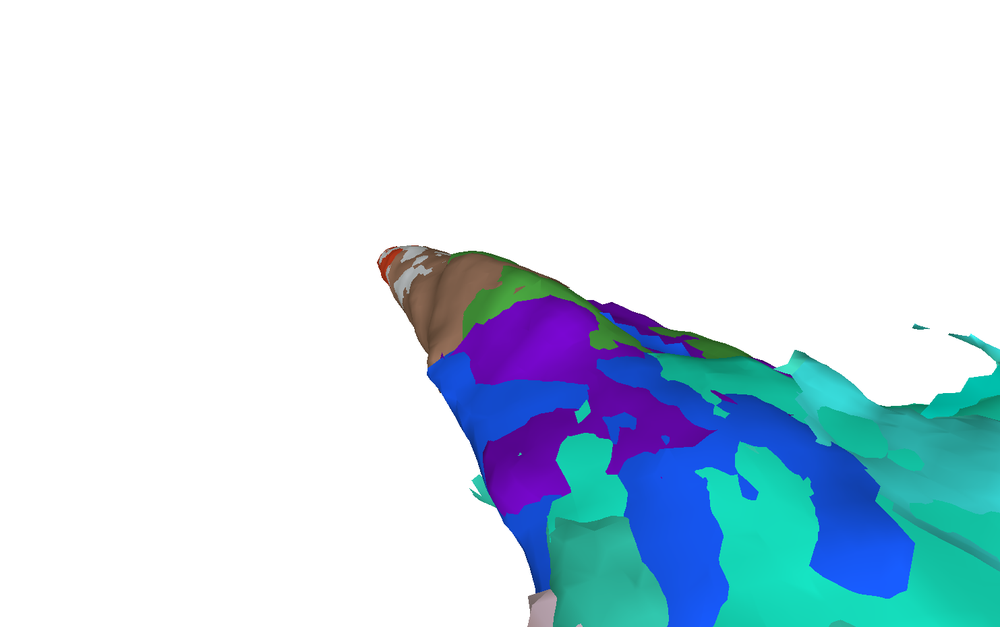}
\end{tabular}

\caption{Mixture Reconstruction Loss. Imposing the reconstruction loss on the mixture instead of directly on the patches leads to individual patches not matching the surface.}
\label{fig:mixture}
\end{figure}

We run an ablation study of each of the extrinsics losses.
We also test whether guiding the rotation via initialization and a loss function helps.
Table~\ref{tab:ablationpatch} contains the results.
Due to our initialization, as described in Sec.~\ref{sec:initialization}, the extrinsics losses are not necessary in this setting.
However, as shown in Sec.~\ref{sec:object} in this supplementary material, they are necessary when the extrinsics are regressed instead of free.
Initializing and encouraging the rotation towards normal alignment helps.
We do not use $\mathcal{L}_\text{recon}$ on the mixture because that modification does not sufficiently constrain the patches to individually reconstruct the surface, as Fig.~\ref{fig:mixture} shows.

\begin{table}[]
\centering
\caption{Ablation Study of PatchNet. We remove each of the extrinsics losses. We also impose the reconstruction loss on the mixture (using $g_i(\mathbf{x})$ from Eq.~11 instead of $f(\mathbf{x},\mathbf{z}_{i,p},\theta)$).
}
\setlength\tabcolsep{3pt} %
\begin{tabular}{l|c|c|c}
\cline{2-4}
                                   & IoU  & Chamfer & F-score \\\hline
no $\mathcal{L}_\text{sur}$       & 92.0 & 0.049   & 94.8    \\
no $\mathcal{L}_\text{cov}$                       & 90.7 & 0.051   & 93.6    \\
no $\mathcal{L}_\text{rot}$              & 92.5 & 0.043   & 95.4    \\
no $\mathcal{L}_\text{scl}$                    & 91.2 & 0.031   & 94.3    \\
no $\mathcal{L}_\text{var}$                 & 91.6 & 0.045   & 94.4    \\\hline
random rotation initialization and no $\mathcal{L}_\text{rot}$       & 89.0 & 0.048 & 93.1   \\\hline
ours                               & 91.6 & 0.045   & 94.5   \\
ours with $\mathcal{L}_\text{recon}$ on mixture        & 94.0 & 0.026   & 96.8   \\\hline
\end{tabular}
\label{tab:ablationpatch}
\end{table}

\section{Error Metrics}\label{sec:metrics}

Similar to Genova~\etal~\cite{genova2019deep}, we evaluate using IoU, Chamfer distance and F-score. 
We report the mean values across different test sets.

\emph{IoU}: For a given watertight groundtruth mesh, we extract the reconstructed mesh using marching cubes at $128^3$ resolution. 
We then sample $100k$ points uniformly in the bounding box of the GT and check for both the generated mesh and the GT whether each point is inside or outside. 
The final value is the fraction of intersection over union, multiplied by a factor of $100$.
Higher is better.

\emph{Chamfer Distance}: Here, we sample 100k points on the surface of both the groundtruth and the reconstructed mesh. 
We use a kD-tree to compute the closest points from the reconstructed to the groundtruth mesh and vice-versa. 
We then square these distances ($L2$ Chamfer) and sum the averages of each direction. For better readability, we finally multiply by 100. 
Lower is better.

\emph{F-score}: For each shape, we threshold the point-wise distances computed before at 0.01 (all meshes are normalized to a unit cube). 
We then compute the fraction of distances below the threshold, separately for each direction. 
Finally, we take the harmonic mean of both these values and multiply the result by 100.
Higher is better.

In cases where a network does not produce any surface, we set the value of IoU to 0, the Chamfer distance to~100, and the F-score to 0.

\section{Reduced Test Set}\label{sec:reduced}

Our reduced test set on ShapeNet consists of 50 randomly chosen test shapes per category.
Table~\ref{tab:reduced} shows how well the error measures on this reduced test set approximate the error measures on the full test set.

\begin{table}
\centering
\caption{Reduced Test Set vs. Full Test Set. The computed metrics on the reduced test set of ShapeNet are a good approximation of the computed metrics on the full test set. This is an extended version of Table~1 from the main paper. } %
\resizebox{\columnwidth}{!}{
\setlength\tabcolsep{3pt} %
\begin{tabular}{l|cc|cc|cc||cc|cc|cc||cc|cc|cc} 
\hline
 & \multicolumn{6}{c||}{IoU}  & \multicolumn{6}{c||}{Chamfer} & \multicolumn{6}{c}{F-score} \\
                          Category &  \multicolumn{2}{c|}{DeepSDF} & \multicolumn{2}{c|}{Baseline} & \multicolumn{2}{c||}{Ours}      &  \multicolumn{2}{c|}{DeepSDF} & \multicolumn{2}{c|}{Baseline} & \multicolumn{2}{c||}{Ours}  & \multicolumn{2}{c|}{DeepSDF} & \multicolumn{2}{c|}{Baseline} & \multicolumn{2}{c}{Ours} \\\hline
                          & full & red. & full & red. & full & red. & full & red. & full & red. & full & red. & full & red. & full & red. & full & red.\\\hline
airplane                  &  84.9 & 84.0 & 65.3 & 64.2 & 91.1 & 90.7 &        0.012 & 0.023 & 0.077 & 0.084 & 0.004 & 0.006 &     93.0 & 92.3& 72.9 & 71.6 & 97.8 & 97.5 \\
bench                     &  78.3 & 77.1 & 68.0 & 65.7 & 85.4 & 83.7 &        0.021 & 0.015 & 0.065 & 0.043 & 0.006 & 0.006 &     91.2 & 90.4 & 80.6 & 80.1 & 95.7 & 94.9 \\
cabinet                   &  92.2 & 89.1 & 88.8 & 84.8 & 92.9 & 91.6 &        0.033 & 0.027 & 0.055 & 0.047 & 0.110 & 0.119 &     91.6 & 90.3 & 86.4 & 84.3 & 91.2 & 91.8 \\
car                       &  87.9 & 88.4 & 83.6 & 84.3 & 91.7 & 92.6 &        0.049 & 0.057 & 0.070 & 0.074 & 0.049 & 0.050 &     82.2 & 82.1 & 74.5 & 74.4 & 87.7 & 87.8 \\
chair                     &  81.8 & 80.1 & 72.9 & 70.3 & 90.0 & 88.6 &        0.042 & 0.041 & 0.110 & 0.118 & 0.018 & 0.013 &     86.6 & 86.0 & 75.5 & 74.8 & 94.3 & 93.5 \\
display                   &  91.6 & 92.9 & 86.5 & 89.1 & 95.2 & 95.5 &        0.030 & 0.010 & 0.061 & 0.034 & 0.039 & 0.049 &     93.7 & 95.1 & 87.0 & 89.8 & 97.0 & 97.3 \\
lamp                      &  74.9 & 72.3 & 63.0 & 63.4 & 89.6 & 88.0 &        0.566 & 2.121 & 0.438 & 0.257 & 0.055 & 0.063 &     82.5 & 79.9 & 69.4 & 70.1 & 94.9 & 94.0 \\
rifle                     &  79.0 & 78.0 & 68.5 & 66.0 & 93.3 & 93.1 &        0.013 & 0.012 & 0.039 & 0.046 & 0.002 & 0.001 &     90.9 & 90.7 & 82.3 & 80.4 & 99.3 & 99.3 \\
sofa                      &  92.5 & 92.2 & 85.4 & 84.5 & 95.0 & 95.1 &        0.054 & 0.075 & 0.226 & 0.236 & 0.014 & 0.012 &     92.1 & 91.3 & 84.2 & 83.0 & 95.3 & 95.3 \\
speaker                   &  91.9 & 90.5 & 86.7 & 84.9 & 92.7 & 90.8 &        0.050 & 0.060 & 0.094 & 0.121 & 0.243 & 0.242 &     87.6 & 84.7 & 79.4 & 75.7 & 88.5 & 85.1 \\
table                     &  84.2 & 83.4 & 71.9 & 69.5 & 89.4 & 90.3 &        0.074 & 0.043 & 0.156 & 0.169 & 0.018 & 0.017 &     91.1 & 91.5 & 79.2 & 79.1 & 95.0 & 96.1 \\
telephone                 &  96.2 & 96.0 & 95.0 & 94.1 & 98.1 & 98.0 &        0.008 & 0.010 & 0.016 & 0.016 & 0.003 & 0.004 &     97.7 & 97.3 & 96.2 & 94.7 & 99.4 & 99.3 \\
watercraft                &  85.2 & 84.9 & 79.1 & 78.5 & 93.2 & 93.1 &        0.026 & 0.019 & 0.041 & 0.031 & 0.009 & 0.006 &     87.8 & 88.2 & 90.2 & 80.6 & 96.4 & 96.6 \\
\hline
mean                      &  86.2 & 85.3 & 78.1 & 76.9 & 92.1 & 91.6 &        0.075 & 0.193 & 0.111 & 0.098 & 0.044 & 0.045 &     89.9 & 89.2 & 80.6 & 79.9 & 94.8 & 94.5 \\
\hline
\end{tabular}
\label{tab:reduced}}
\end{table}

\section{Generalization Experiments -- Single Shape}\label{sec:single}
\begin{figure}
\centering
\includegraphics[trim={200 0 50 50},clip,height=3cm]
{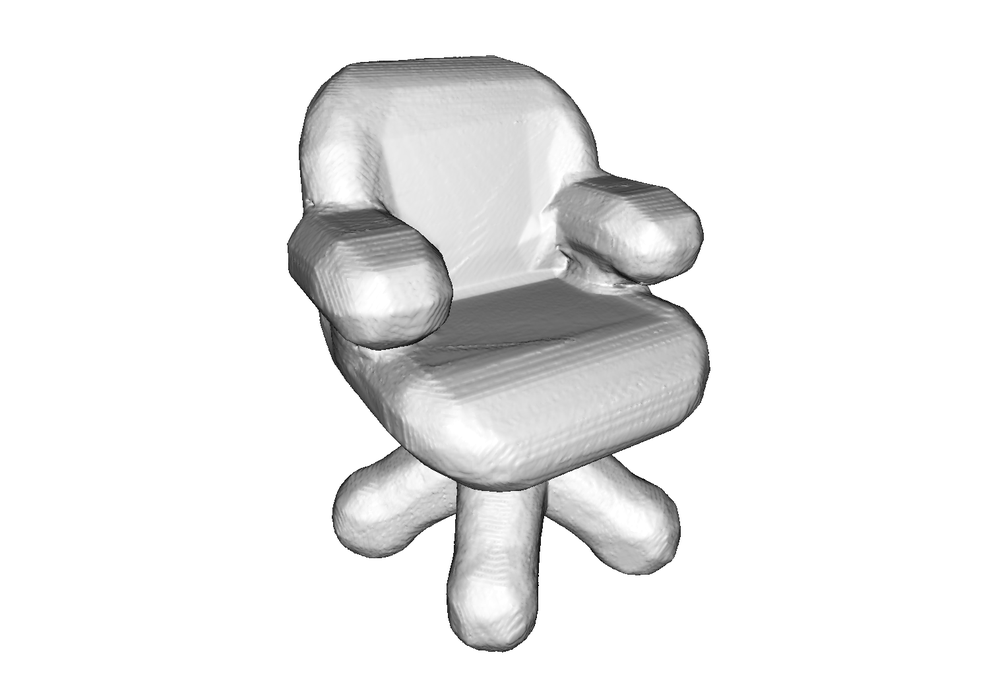}

\caption{Single Shape. In one of the generalization experiments in Sec.~\ref{sec:generalization} of the main paper, we train PatchNet on this randomly chosen groundtruth shape.}
\label{fig:single}
\end{figure}

Fig.~\ref{fig:single} shows the randomly picked single shape on which we trained PatchNet in Sec.~\ref{sec:generalization} of the main paper.

\section{Object-Level Priors}\label{sec:object}

\subsection{Surface Reconstruction}
We report surface reconstruction errors using object-level priors (see Sec.~\ref{sec:objectprior} from the main paper).
Note that the experiments in Sec.~\ref{sec:objectprior} of the main paper use the \emph{most} competitive setting of the global-patch baseline (\emph{i.e.}, pretrained on all categories and then refined) and the \emph{least} competitve setting of PatchNet (\emph{i.e.}, pretrained on one category and not refined).
This demonstrates how well our proposed PatchNet generalizes.
For consistency, for the DeepSDF-based baseline, we choose the same setting as for the global-patch baseline.
Note that that setting is virtually on par with the most competitive DeepSDF setting (\emph{i.e.}, pretrained on one category and then refined).

\subsubsection{Settings}
Both our network and the baselines consist of a four-layer ObjectNet and the standard final eight FC layers.
We pretrain the final eight FC layers either on the reduced training set of all categories or on all shapes from the \emph{Cabinets} category training set.
We then either keep those pretrained weights fixed while training ObjectNet or we allow them to be refined.
While at training time, each phase lasts 1000 epochs, we reduce this to 800 epochs at test time.

\subsubsection{Results}

Table~\ref{tab:objectsurface} contains the quantitative results.
The baselines do not generalize well if they are kept fixed.
Refinement improves error measures.

\newcommand{\STAB}[1]{\begin{tabular}{@{}c@{}}#1\end{tabular}}
\begin{table}
\centering
\caption{Surface Reconstruction with ObjectNet. We pretrain the final eight layers either on one category (\emph{one}) or on all categories (\emph{all}). We then either keep those layers fixed (\emph{fix.}) or refine them (\emph{ref.}).}
\setlength\tabcolsep{4pt} %
\resizebox{\columnwidth}{!}{
\begin{tabular}{cl|cc|cc|cc|cc|cc|cc}
\cline{3-14}
&          & \multicolumn{4}{c|}{baseline} & \multicolumn{4}{c|}{DeepSDF-based} & \multicolumn{4}{c}{ours} \\ \cline{3-14}
&          & \multicolumn{2}{c|}{one} & \multicolumn{2}{c|}{all} & \multicolumn{2}{c|}{one} & \multicolumn{2}{c|}{all} & \multicolumn{2}{c|}{one} & \multicolumn{2}{c}{all} \\ \cline{3-14}
&          & fix. & ref. & fix. & ref. & fix. & ref. & fix. & ref. & fix. & ref. & fix. & ref.\\\hline
 \multirow{3}{*}{\STAB{\rotatebox[origin=c]{90}{\tiny airplanes}}} 
& IoU      & 35.9 & 70.9 & 60.2 & 73.3 & 47.0 & 75.6 & 69.9 & 74.1 & 67.5 & 68.5 & 71.9 & 74.2 \\
& Chamfer  & 0.710& 0.146& 0.218& 0.147& 0.546& 0.049& 0.127& 0.050& 0.203& 0.182& 0.179& 0.170\\
& F-score  & 37.5 & 76.0 & 63.6 & 78.3 & 49.1 & 82.5 & 76.4 & 81.6 & 71.7 & 74.1 & 77.9 & 79.7 \\\hline
\multirow{3}{*}{\STAB{\rotatebox[origin=c]{90}{\tiny sofas}}}     
& IoU      & 76.1 & 81.8 & 76.3 & 84.3 & 76.4 & 79.7 & 82.4 & 76.6 & 85.3 & 86.2 & 84.9 & 86.0 \\
& Chamfer  & 0.416& 0.159& 0.398& 0.171& 0.467& 0.178& 0.282& 0.406& 0.118& 0.139& 0.236& 0.082\\
& F-score  & 69.0 & 75.2 & 71.8 & 77.9 & 70.1 & 72.3 & 77.5 & 71.8 & 79.0 & 80.7 & 79.5 & 79.9 \\\hline
\end{tabular}
}
\label{tab:objectsurface}
\end{table}

\subsection{Ablation Study}
We evaluate the extrinsics losses in the context of surface reconstruction with object-level priors.
We use the version of our method from the main paper: pretrained on the \emph{Cabinets} category and without refinement. 
We perform the ablation study on the \emph{Sofas} category.

The quantitative results are in Table~\ref{tab:ablationobject}.
The network failed to reconstruct without $\mathcal{L}_\text{cov}$.

\begin{table}[]
\centering
\caption{Ablation Study with Object-level Priors. We remove each of the extrinsics losses.}
\setlength\tabcolsep{3pt} %
\begin{tabular}{l|c|c|c}
\cline{2-4}
                                               & IoU  & Chamfer & F-score \\\hline
no $\mathcal{L}_\text{sur}$                    & 87.6 & 0.076 & 82.6  \\
no $\mathcal{L}_\text{scl}$                    & 75.5 & 0.154 & 54.2    \\
no $\mathcal{L}_\text{var}$                    & 71.8 & 0.269 & 47.3   \\\hline
ours                                           & 85.3 & 0.118 & 79.0   \\
ours with $\mathcal{L}_\text{recon}$ on mixture  & 84.9 & 0.116 & 78.1  \\ \hline
\end{tabular}
\label{tab:ablationobject}
\end{table}

\subsection{Partial Point Cloud Completion}
We report additional depth-map completion results using the same settings for our method that we use for the baselines in the main paper (pretrained on all categories and refined).
Note that in the main paper, we report the shape-completion results of the most disadvantageous version of our method (according to Table~\ref{tab:objectsurface}).
Table~\ref{tab:shape_completion} contains the quantitative results.
In all cases, our method after local refinement yields the best results.

\begin{table}
\centering
\caption{Partial Point Cloud Completion from Depth Maps. We complete depth maps from a fixed camera viewpoint and from per-scene random viewpoints.}
\setlength\tabcolsep{3pt} %
\resizebox{\columnwidth}{!}{
\begin{tabular}{l|c|c|c|c|c|c|c|c}
\cline{2-9}
             & \multicolumn{2}{c|}{sofas fixed}      & \multicolumn{2}{c|}{sofas random}     & \multicolumn{2}{c|}{airplanes fixed}      & \multicolumn{2}{c}{airplanes random}     \\
             & acc.       & F-score       & acc.       & F-score       & acc.       & F-score       & acc.       & F-score       \\\hline
baseline                        & 0.094 & 43.0 & 0.092 & 42.7 & 0.069 & 58.1 & 0.066 & 58.7 \\
DeepSDF-based baseline          & 0.106 & 33.6 & 0.101 & 39.5 & 0.066 & 56.9 & 0.065 & 55.5 \\
ours (main paper)               & 0.091 & 48.1 & 0.077 & 49.2 & 0.058 & 60.5 & 0.056 & 59.4 \\
ours+refined (main paper)       & \textbf{0.052} & 53.6 & \textbf{0.053} & 52.4 & \textbf{0.041} & 67.7 & \textbf{0.043} & 65.8 \\
ours (baseline-matched)         & 0.088 & 47.5 & 0.074 & 50.0 & 0.052 & 64.8 & 0.050 & 64.3 \\
ours+refined (baseline-matched) & 0.061 & \textbf{54.7} & 0.056 & \textbf{53.5} & 0.045 & \textbf{70.3} & 0.044 & \textbf{69.9} \\        \hline 
\end{tabular}}
\label{tab:shape_completion}
\end{table}

\section{Number of Patches and Network/Latent Code Sizes}\label{sec:patches}

\begin{figure}
\centering
\includegraphics[trim={0 0 65 0},clip,height=3.6cm]
{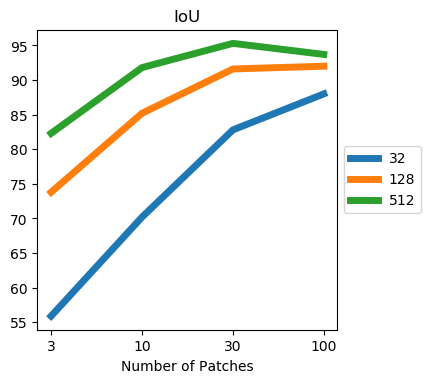}
\hfill
\includegraphics[trim={0 0 65 0},clip,height=3.6cm]
{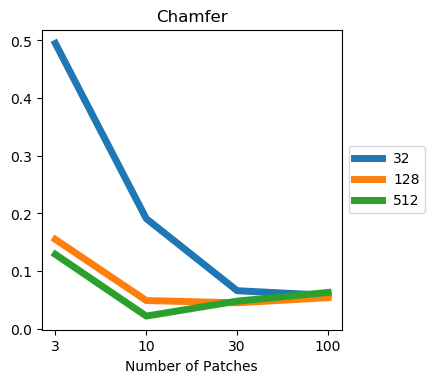}
\hfill
\includegraphics[trim={0 0 0 0},clip,height=3.6cm]
{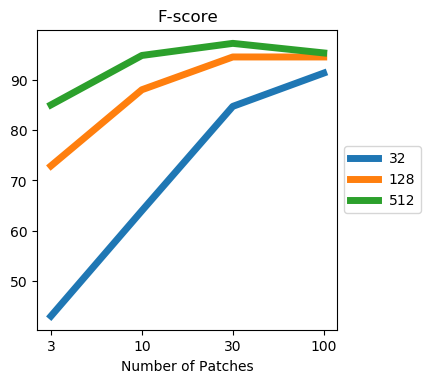}
\caption{Mean error metrics on the reduced ShapeNet test set for different numbers of patches and network/latent code sizes. }
\label{fig:sizes}
\end{figure}

Fig.~\ref{fig:sizes} shows the mean error metrics on the reduced ShapeNet test set when training on the reduced ShapeNet training set.
We try out different sizes. Size refers to both the dimensions of the patch latent vector and the hidden dimensions of PatchNet, as in Sec.~\ref{sec:main_ablation}.
The gap between size 128 and 512 is much smaller than between 32 and 128.
Furthermore, using 100 patches instead of 30 yields only marginal gains at best.

\begin{figure}
\centering
\includegraphics[trim={0 0 100 0},clip,height=3.6cm]
{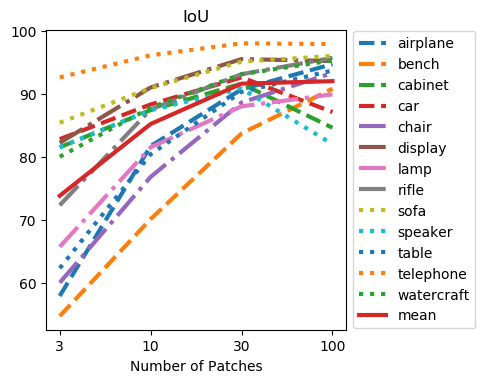}
\hfill
\includegraphics[trim={0 0 95 0},clip,height=3.6cm]
{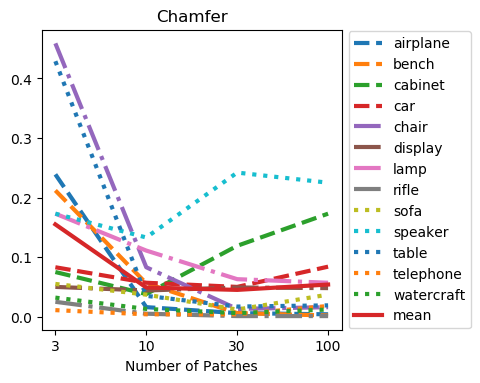}
\hfill
\includegraphics[trim={0 0 0 0},clip,height=3.6cm]
{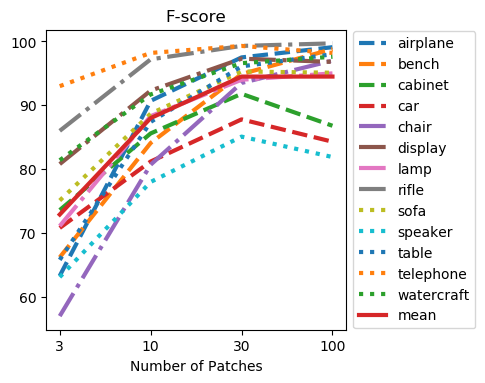}

\caption{Per-category error metrics on the reduced ShapeNet test set for different numbers of patches.}
\label{fig:categories}
\end{figure}

Fig.~\ref{fig:categories} shows the per-category error metric on the reduced ShapeNet test set when training on the reduced ShapeNet training set.
We conduct this experiment with different numbers of patches.
Apart from the outlier categories \emph{cabinet}, \emph{car}, and \emph{speaker}, we observe that the error metrics behave very similar across categories.
They improve strongly when going from 3 to 10 and from 10 to 30 patches and they improve at most slightly when going from 30 to 100 patches.

\section{Synthetic Noise}\label{sec:noise}

We investigate the robustness of PatchNet by adding Gaussian noise to the groundtruth SDF values of the reduced test set.
We use the PatchNet trained with default settings, which also means that it has only seen unperturbed SDF data during training.
The Gaussian noise has zero mean and different standard deviations $\sigma$.
For reference, the mesh fits tightly into the unit sphere, as mentioned in Sec.~\ref{sec:prelim}.
The results are in Table~\ref{tab:noise}.

\begin{table}[]
\centering
\caption{Synthetic Noise at Test Time.}
\setlength\tabcolsep{3pt} %
\begin{tabular}{l|c|c|c}
\cline{2-4}
                                               & IoU  & Chamfer & F-score \\\hline
$\sigma=0.1$   & 81.2 & 0.037 & 85.3 \\ 
$\sigma=0.01$  & 90.3 & 0.045 & 94.3 \\ 
$\sigma=0.001$ & 91.5 & 0.047 & 94.4 \\
\hline
$\sigma=0$ (ours) & 91.6 & 0.045   & 94.5   \\
\end{tabular}
\label{tab:noise}
\end{table}

\section{Preliminary Results on ICL-NUIM}\label{sec:large}

Once trained, a PatchNet can be used for any number of patches at test time.
Here, we present some preliminary results on the large living room from ICL-NUIM \cite{handa:etal:ICRA2014}.

Since the scene is already watertight, we skip the depth fusion step of the preprocessing method.
We reduce the standard deviation used to generate SDF samples by a factor of 100 to account for scaling differences.
Overall, we sample 50 million SDF samples.

For PatchNet, we use 800 patches.
We keep the extrinsics fixed at their initial values since we found that to improve the reconstruction.
We optimize for 10k iterations, halving the learning rate every 2k iterations.
During optimization, 25k SDF samples are used per iteration.
The baselines are trained with the same modified settings.

\begin{figure}
\centering

\includegraphics[trim={0 0 0 0},clip,height=3.5cm]
{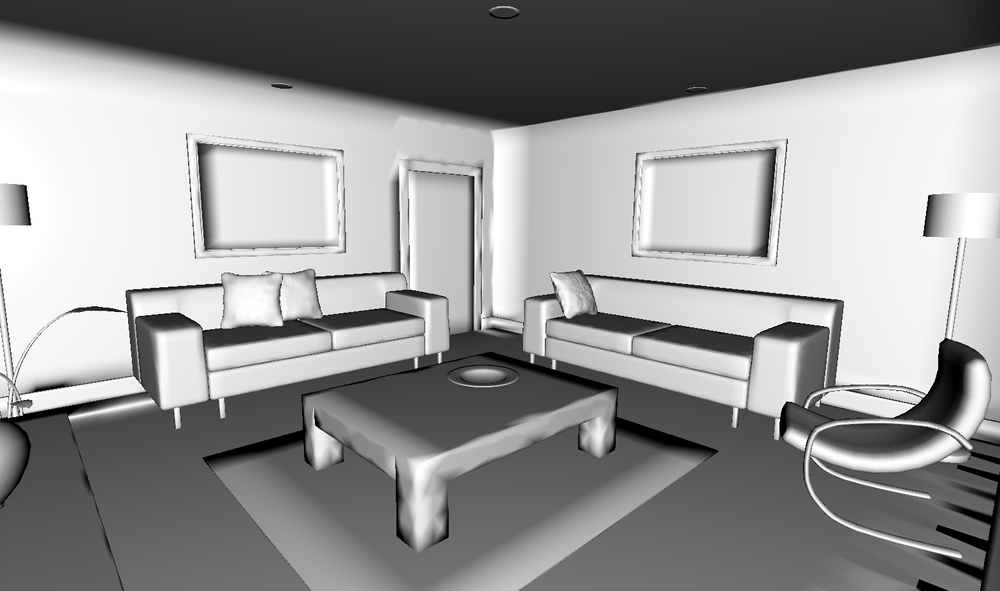}
\\
{\centering Groundtruth}%
\\
\includegraphics[trim={0 0 0 0},clip,height=3.5cm]
{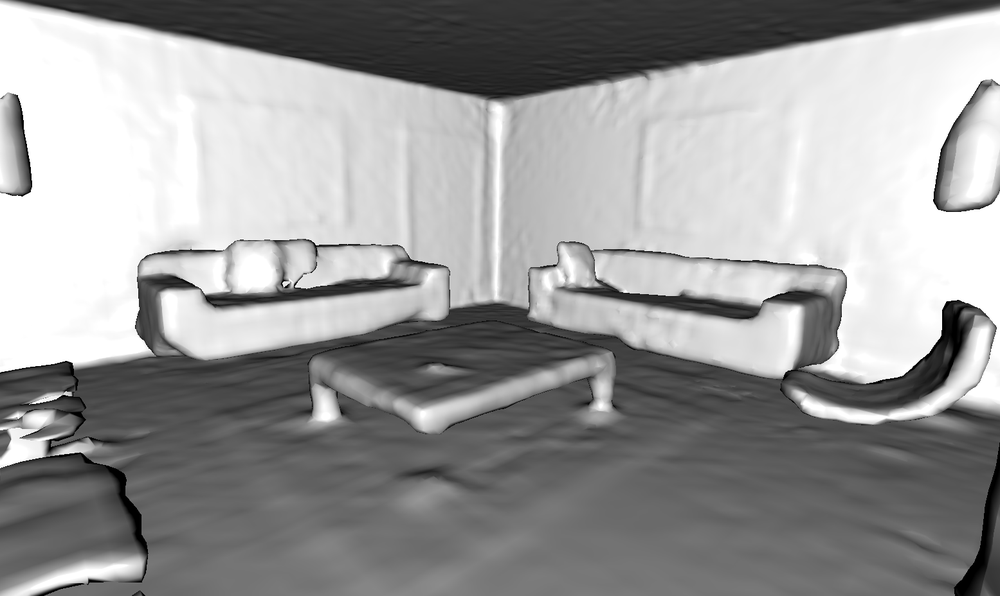}
\hfill
\includegraphics[trim={0 0 0 0},clip,height=3.5cm]
{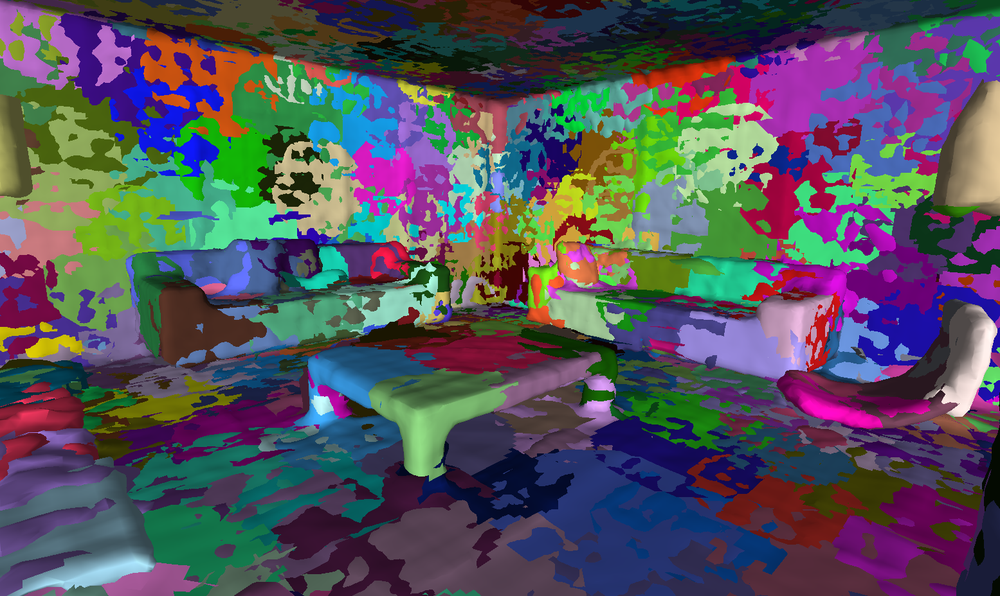}
\\
{\centering Mixture (Ours) \hspace{12em} Patches (Ours) }%
\\
\includegraphics[trim={0 0 0 0},clip,height=3.5cm]
{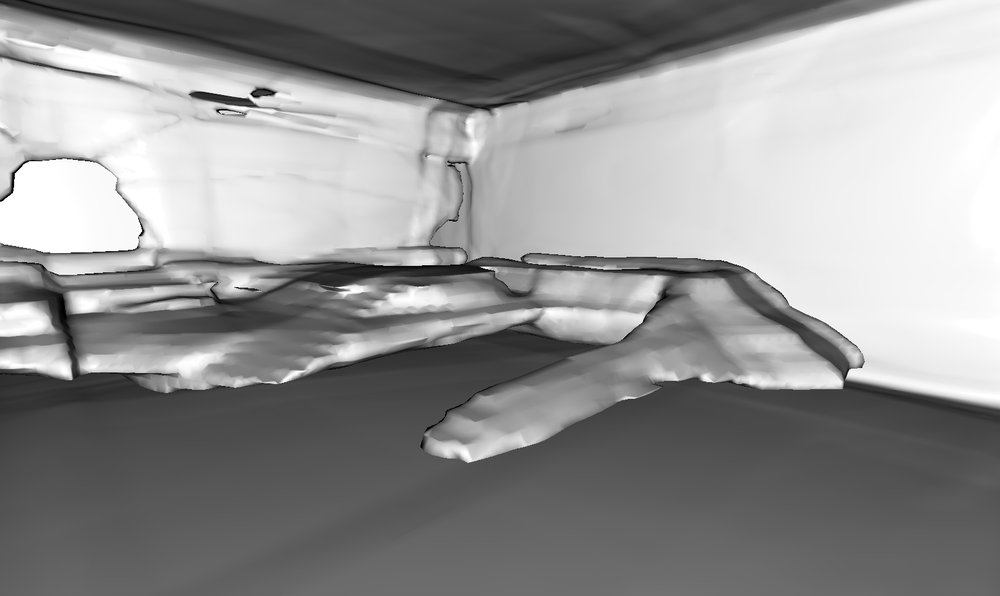}
\hfill
\includegraphics[trim={0 0 0 0},clip,height=3.5cm]
{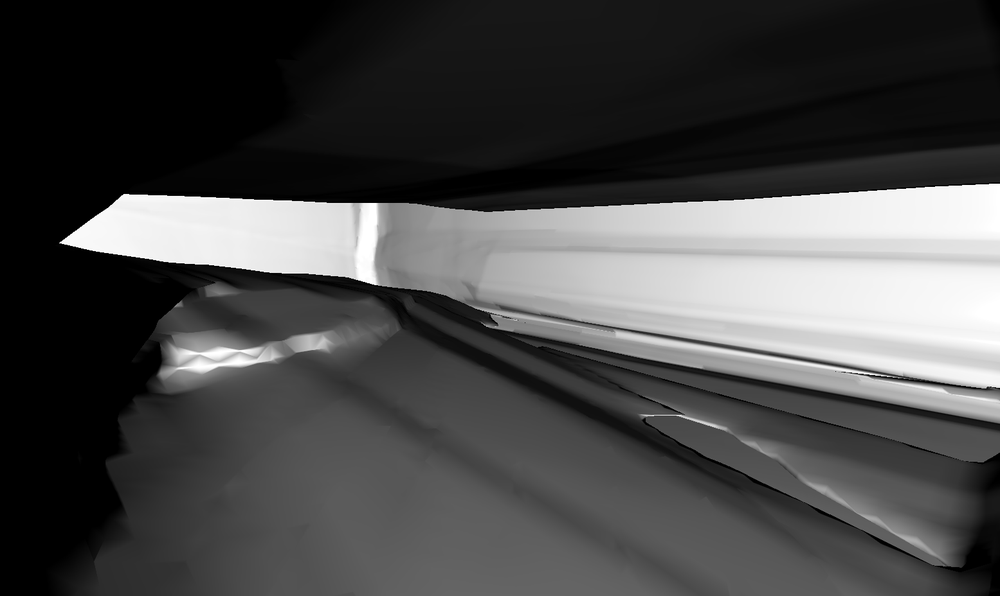}
\\
{\hspace{7em} DeepSDF \hspace{12em} Our Global Baseline \hfill}

\caption{Preliminary Results on ICL-NUIM.}
\label{fig:large}
\end{figure}

The results are in Fig.~\ref{fig:large}.
Note that due to our extrinsics initialization (Sec.~\ref{sec:prelim}) and $\mathcal{L}_\text{var}$, all patches have similar sizes, which leads to a wasteful distribution.

\section{Remarks on the Concurrent Work DSIF \cite{genova2019deep}}\label{sec:dsif}

For completeness, we provide some remarks on the unpublished, but concurrent related work \emph{Deep Structured Implicit Functions (DSIF)} by Genova~\emph{et al.}~\cite{genova2019deep}\footnote{After submission of this work, DSIF was published at CVPR 2020 and renamed to \emph{Local Deep Implicit Functions for 3D Shape}.}.
In our terminology, they use a network from prior work (SIF~\cite{genova2019learning}) to regress patch extrinsics from depth maps of 20 fixed viewpoints.
They then use a point-set encoder to regress patch latent codes from backprojected depth maps according to the regressed extrinsics.
Finally, they propose a modified version of OccupancyNetworks~\cite{mescheder2019occupancy} to regress point-wise occupancy probabilities.

As Table~2 in the main paper shows, our proposed method outperforms theirs almost everywhere despite being trained on only $\sim4\%$ of the training data.
Since they impose their reconstruction loss on the final mixture, we do the same for a comparison in Table~\ref{tab:ablationpatch} in this supplementary material.
Using 32 patches and $N_z=128$, their method obtains an F-score below $95$ (on the full test set), while our method reaches $96.8$ (on the reduced test set; which is very representative of the full test set, see Sec.~\ref{sec:reduced}).

Furthermore, they regress the patch extrinsics with a network taken from prior work~\cite{genova2019learning}, while we show that it is possible to directly and effectively initialize them.
Because DSIF regresses extrinsics, it can have issues predicting extrinsics for shapes very different from the training data, while we \emph{by construction} do not have such issues.
It also turns out that the isotropic Gaussian weights we use in our proposed method are sufficient to outperform their method, which uses more complicated anisotropic Gaussians.
Finally, for their encoder to work, the input geometry needs to be represented in some way (which is a non-trivial decision that might impact performance), while we avoid this issue by auto-decoding.%

\end{document}